\documentclass[lettersize,journal]{IEEEtran}
\usepackage{amsmath,amsfonts}
\usepackage{algorithmic}
\usepackage{array}
\usepackage[labelformat=parens,labelsep=space]{subcaption}
\usepackage{textcomp}
\usepackage{stfloats}
\usepackage{url}

\usepackage{hyperref}
\hypersetup{
    colorlinks=false,       
    linkcolor=blue,         
    filecolor=magenta,      
    urlcolor=cyan,          
    }
    
\usepackage{verbatim}
\usepackage{graphicx}
\usepackage{multirow}
\usepackage[inkscapelatex=false]{svg}
\usepackage{hyphenat}  
\usepackage{diagbox}
\usepackage{footnote}

\usepackage{tabularray}

\usepackage[nolist,nohyperlinks]{acronym}
\newacro{KG}{Knowledge Graph}
\newacro{KGE}{Knowledge Graph Embedding}
\newacro{TTC}{Time To Collision}
\newacro{CNN}{Convolutional Neural Network}
\newacro{VRU}{Vulnerable Road User}
\newacro{NHTSA}{National Highway Traffic Safety Administration}
\newacro{AV}{Autonomous Vehicle}
\newacro{LSTM}{Long Short-Term Memory}
\newacro{KG}{Knowledge Graph}
\newacro{KGE}{Knowledge Graph Embedding}
\newacro{JAAD}{Joint Attention for Autonomous Driving}
\newacro{PSI}{Pedestrian Situated Intent}
\newacro{RAG}{Retrieval Augmented Generation}
\newacro{LLM}{Large Language Model}
\newacro{MRR}{Mean Reciprocal Rank}
\newacro{C3D}{Convolutional 3D}
\newacro{fc}{fully-connected}
\newacro{PCPA}{Pedestrian crossing prediction with attention}
\newacro{TS}{Takagi-Sugeno}
\newacro{eP2P}{Pedestrian Trajectory Prediction}
\newacro{MRR}{Mean Reciprocal Rank}

\usepackage{cleveref}

\begin{document}
\title{RAG-based Explainable Prediction of Road Users Behaviors for Automated Driving using Knowledge Graphs and Large Language Models}

\author{Mohamed Manzour Hussien, Angie Nataly Melo, Augusto Luis Ballardini, Carlota Salinas Maldonado, Rubén Izquierdo and Miguel Ángel Sotelo \IEEEmembership{Fellow, IEEE}
\thanks{
All authors are with the Computer Engineering Department, University of Alcalá, Alcalá de Henares, 
Madrid, Spain. Email: 
\href{mailto://ahmed.manzour@uah.es}{ahmed.manzour@uah.es},
\href{mailto://nataly.melo@uah.es}{nataly.melo@uah.es},
\href{mailto://augusto.ballardini@uah.es}{augusto.ballardini@uah.es},
\href{mailto://carlota.salinasmaldo@uah.es}{carlota.salinasmaldo@uah.es},
\href{mailto://ruben.izquierdo@uah.es}{ruben.izquierdo@uah.es},
\href{mailto://miguel.sotelo@uah.es}{miguel.sotelo@uah.es}
}}


\maketitle

\begin{abstract}
Prediction of road users’ behaviors in the context of autonomous driving has gained considerable attention by the scientific community in the last years. Most works focus on predicting behaviors based on kinematic information alone, a simplification of the reality since road users are humans, and as such they are highly influenced by their surrounding context. In addition, a large plethora of research works rely on powerful Deep Learning techniques, which exhibit high performance metrics in prediction tasks but may lack the ability to fully understand and exploit the contextual semantic information contained in the road scene, not to mention their inability to provide explainable predictions that can be understood by humans. In this work, we propose an explainable road users’ behavior prediction system that integrates the reasoning abilities of Knowledge Graphs (KG) and the expressiveness capabilities of Large Language Models (LLM) by using Retrieval Augmented Generation (RAG) techniques. For that purpose, Knowledge Graph Embeddings (KGE) and Bayesian inference are combined to allow the deployment of a fully inductive reasoning system that enables the issuing of predictions that rely on legacy information contained in the graph as well as on current evidence gathered in real time by onboard sensors. Two use cases have been implemented following the proposed approach: 1) Prediction of pedestrians’ crossing actions; 2) Prediction of lane change maneuvers. In both cases, the performance attained surpasses the current state of the art in terms of anticipation and F1-score, showing a promising avenue for future research in this field.
\end{abstract}


\section{Introduction and related work}
\label{sec:introductoin-and-related-work}
\begin{figure}
  \centering   
    \includegraphics[width=1.00\columnwidth]{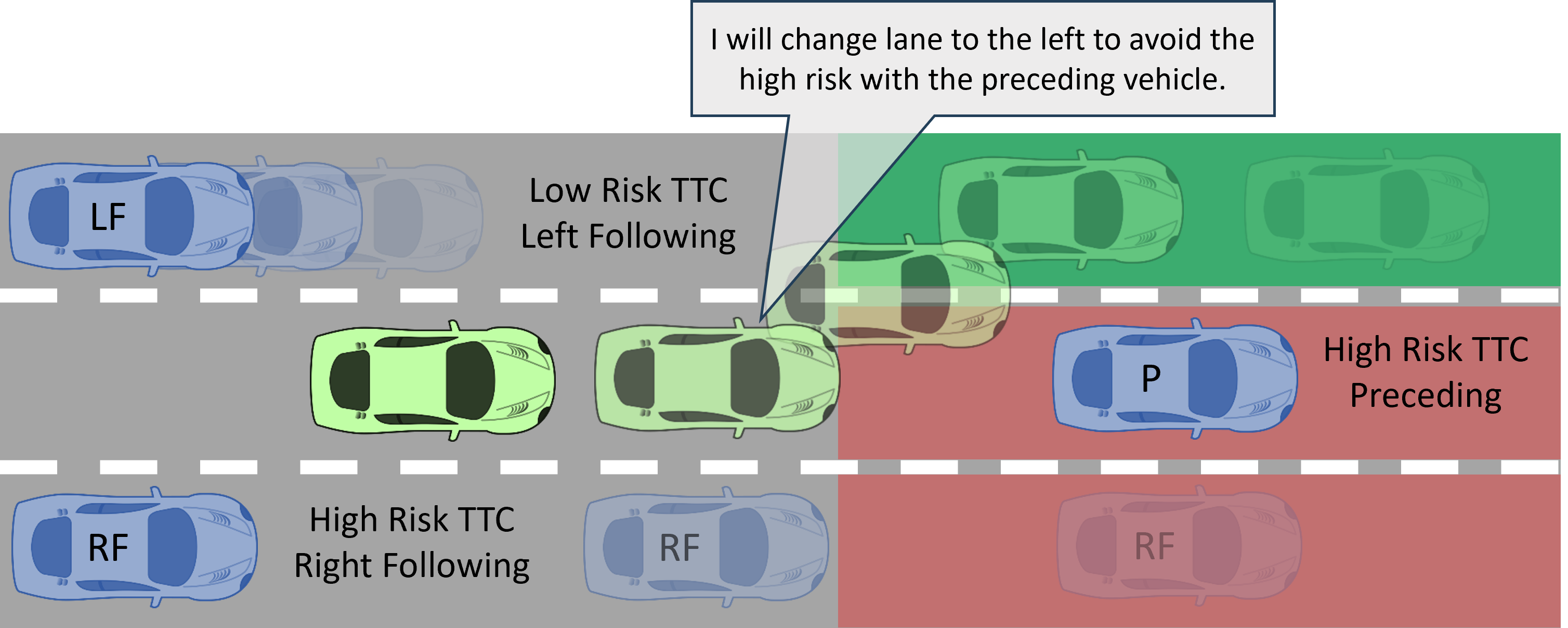}      
    \caption{The target vehicle (green) will most likely make a left lane change maneuver based on the risk assessment of the surrounding (blue) vehicles.}
  \label{fig:figure-manzour}
\end{figure}

\IEEEPARstart{D}{espite} the significant progress that the world has experienced in the last years in terms of road safety, road traffic deaths continue to represent a global health crisis, according to the World Organization report on road safety \cite{WHO}, especially for \acp{VRU}, which are involved in 53\% of all road traffic fatalities. The same report highlights the fact that 23\% of fatal accidents involve pedestrians. As a matter of act, pedestrians are the most vulnerable road user group also on European Union roads, being involved in 20\% of road traffic fatalities \cite{Freya}. Similarly, the statistics published in 2023 by the \ac{NHTSA} reveal an increase in the number of deaths in motor vehicle traffic crashes in the United States of America in 2021 as compared to 2020, and a 17.3\% increase compared to 2019 \cite{Stewart}, being lane-changing maneuvers one of the main causes for vehicle crashes, as the same report indicates that 33\% of all road crashes take place during a lane change maneuver. These figures advocate for the need to develop technologies aiming at enhancing road safety by endowing automated vehicles with the capacity to anticipate pedestrians’ and drivers’ behaviors and motion patterns, such as road crossing actions (for pedestrians) and lane change maneuvers (for drivers). The ability to characterize and predict the behavior and motion patterns of road users, namely drivers and vulnerable road users (pedestrians and cyclists), as well as the explanation and understanding of the factors that rule the interactions among them, is essential for increasing road safety and traffic efficiency in the context of autonomous driving. Furthermore, the possibility of deploying \acp{AV} with the capability of understanding and anticipating road users’ behaviors will also allow to increase the perception of comfort and the feeling of “being understood and respected” in all road users interacting with \acp{AV} (pedestrians, cyclists, drivers of manually driven vehicles), a feature that will definitely contribute to the social acceptance of \acp{AV} and, consequently, to accelerate their commercial deployment.   
In the context of Autonomous Driving, a large plethora of road users’ behavior estimation algorithms have been developed to predict forthcoming actions of pedestrians \cite{Tsotsos1}, cyclists \cite{Pool}, and drivers \cite{Izquierdo1}, in an attempt to understand and anticipate their behaviors. However, there is a missing component in the literature, namely a holistic view of road users’ behavior and decision making to identify the extent of factors that affect their behaviors and to explain in what ways they are interrelated. This is due to the fact that the majority of research works in the literature disregard the theoretical findings of traffic interaction and treat the problem as dealing with rigid dynamic objects rather than a social being \cite{Schulz}. The behavior of road users depends on a considerable number of factors, such as human factors, gender and age, awareness level or gaze direction, etc. For pedestrians and cyclists, their head orientation and fully articulated body pose provide cues of great relevance for prediction purpose. Something similar can be said about the state of blinkers or braking lights in the case of vehicles. A second set of factors accounts for the influence of other road users, that can be considered individually or in groups (especially, at cross walks). Third, some environmental factors are of utmost relevance, such as the road topology structure and street layout, weather and lighting conditions, traffic rules, signalization, state and type of pavement, etc. On top of that, there are cross-cultural differences that transversally affect the three aforementioned types of factors, i.e., road users may behave differently in the same road scenario depending on the region or country, due to different social norms. This rationale suggests the need for incorporating contextual information when developing road users’ behavior understanding and prediction systems, where the context can include information of different nature, such as kinematic, body language, attention, gaze direction, traffic status per lane, road layout and conditions, and, even, social norms. In other words, context, in a holistic sense, is key to behavior understanding. As an example, \Cref{fig:figure-manzour} depicts a situation involving several vehicles driving on a three-lane highway. The green vehicle, which drives along the middle lane, represents the vehicle under interest, also referred to as target vehicle, while the blue vehicles represent the vehicles around the target vehicle, being denoted as Left Following (LF, i.e., the vehicle driving behind the target vehicle on the left adjoining lane), Right Following (RF, i.e., the vehicle driving behind the target vehicle on the right adjoining lane), and Preceding (P, i.e., the vehicle preceding the target vehicle along its ego-lane, which in this example is the middle lane), respectively. The figure shows the situation at the current time (in solid colors) as well as several future positions of all vehicles (in semi-transparent colors) according to the most likely predictions. In this scenario, the target vehicle is approaching quickly and dangerously the preceding vehicle due to a high difference in velocity between them, thus there is a high risk of collision with P based on the analysis of the estimated time to collision (TTC). Similarly, the LF vehicle is driving even faster than the target vehicle, making a Left lane change maneuver not advisable due to the high risk of collision with LF in such case. In this situation the target vehicle has two possible courses of action: 1) stay on the same lane while decreasing velocity abruptly in order to avoid a collision with P; 2) stay on the same lane and postpone the left lane change maneuver, which seems to be the safer maneuver for all the actors involved in the scene. Action two (left lane change) appears to be a natural behavior that can be executed in a soft, organic manner in coordination with the three surrounding vehicles. Consequently, the behavior understanding and prediction system of LF would regard the left lane change maneuver of the target vehicle as a very likely future behavior and would anticipate and prepare its ego-actions accordingly in order to accommodate such a potential maneuver of the target vehicle (most likely by decreasing speed gently to make the left lane change maneuver of the target vehicle even smoother and safer for all the actors involved). Accounting for contextual information is essential for understanding behavior in this situation. This context-based reasoning approach can be extended and applied to many other similar situations, involving pedestrians and drivers, where contextual information is key for understanding and anticipating behavior. For example, let’s consider an urban avenue with two-lanes per direction of travel where the ego-vehicle drives on the left most lane and tries to overtake a car that drives on the right most lane. Suddenly, the car on the right lane decelerates abruptly while approaching a pedestrian crossing. From the position of the ego-vehicle no pedestrian is visible, thus, the maneuver of the other car may seem unjustified. However, when the ego-vehicle approaches the pedestrian crossing, a pedestrian (previously occluded by the car on the right most lane or by parked cars) becomes fully visible. Anticipating the presence of the occluded pedestrian can best be done by leveraging the semantic understanding of the situation.

A large number of research works have dealt with road users’ behavior prediction in the context of autonomous driving. Regarding the prediction of lane change maneuvers, in \cite{Su} the authors used a \ac{LSTM} model to predict vehicle lane changes by considering the vehicle’s past trajectory and neighbors’ states. In \cite{Benterki} two machine learning models were utilized to predict lane changes of surrounding vehicles on highways. The inputs were longitudinal/lateral velocities, longitudinal/lateral accelerations, distance to left/right lane markings, yaw angle, and yaw rate related to the road. These inputs were trained and tested on Support Vector Machines (SVM) and Artificial Neural Networks (ANN) models. In \cite{Izquierdo2} the authors predicted lane change intentions of surrounding vehicles using two different methodologies and by only considering the visual information provided by the PREVENTION dataset \cite{Izquierdo3}. The first method was Motion History Image - Convolutional Neural Network (MHI-CNN), where temporal and visual information was obtained from the MHI, and then fed to the CNN model. The second model was the GoogleNet-LSTM model, in which a feature vector was obtained from a GoogleNet CNN model and then fed to the \ac{LSTM} model to learn temporal patterns. In \cite{Laimona} the authors trained \ac{LSTM} and Recurrent Neural Networks (RNN) models on the PREVENTION dataset to predict surrounding vehicles’ lane changing intentions by tracking the vehicles’ positions (centroid of the bounding box). Sequences of 10, 20, 30, 40, and 50 frames of (X, Y) coordinates of the target vehicle were considered for comparison. It was concluded that RNN models performed better on short sequence lengths and the \ac{LSTM} model outperformed RNN at long sequences. The work implemented in \cite{Xue} utilized eXtreme Gradient Boosting (XGBoost) and \ac{LSTM} to predict the vehicle lane change decision and trajectory prediction, respectively, in scenarios in the HighD dataset \cite{Krajweski}. The models were based on the traffic flow (traffic density) level, the type of vehicle, and the relative trajectory between the target vehicle and surrounding vehicles. Vehicle trajectory predictions were issued based on historical trajectories and the predicted lane change decisions. In \cite{Gao} a dual Transformer model was proposed. The first Transformer was intended for lane change prediction, while the second one was used for trajectory prediction. Similarly, the prediction of pedestrians’ crossing actions is another task that has been intensively targeted by the research community, focusing on forecasting whether or not a target pedestrian will cross the road at some point in the near future (typically in the next 1-5 seconds). This task has been addressed through a diverse range of algorithms and architectures. Among these approaches, it is particularly noteworthy to highlight a number of methods as SingleRNN based on Recurrent Neural Networks (RNNs) \cite{Tsotsos2}, CapFormer which uses a self-attention alternative based on transformer architecture \cite{Lorenzo}, a 3D Convolutional model (C3D) based on spatiotemporal feature learning \cite{Tran}, a stacked multilevel fusion RNN (SFRNN) \cite{Tsotsos3}, and convolutional \ac{LSTM} (ConvLSTM) \cite{Xingjian}. Despite the abundance of models and research focused on pedestrian crossing predictions, only a limited number of them provide insights into explainability or are specifically developed within the context of explainability. For instance, the research \cite{Achaji} highlights that Transformers offer an advantage in terms of interpretability, due to their attention mechanism. Moreover, the utilization of pedestrian location and body keypoints as features in predicting pedestrian actions results in more human-like behavior. In \cite{Muscholl}, the authors propose a dynamic Bayesian network model that takes into account the influence of interaction and social signals. This system leverages visual means and employs various inference methods to provide explanations for its predictions, with a specific focus on determining the relative importance of each feature in influencing the probability of pedestrian actions.

While Deep Learning (DL) techniques have been reasonably successful in solving road users’ behavior prediction tasks, they may lack the ability to fully understand and exploit the interdependences between road users and the semantic relations implicit in a road scene. Not in vain, in real world applications it is impractical and inefficient to learn all facts and data patterns from scratch, especially when prior and linguistic knowledge is available. As an alternative, neuro-symbolic learning \cite{Yi} has the capacity to exploit such information to further improve the ability to really understand road scenes by utilizing well-formed axioms and rules that can guarantee explainability, both in terms of asserted and inferred knowledge. In neuro-symbolic systems, abstract knowledge extraction is first carried out by means of neural Deep Learning (DL) techniques, that transform the reality into symbols, while logic (or symbolic) reasoning is then performed on the grounds of such symbols. This human-like reasoning approach is interpretable and disentangled, while allowing for compositional, accurate, and generalizable reasoning in rich, complex contexts, such as road scene understanding and autonomous driving, that require identifying and reasoning about entities (road users, road context, and events) that are bundled together by means of spatial, temporal, social, and semantic relations. Knowledge infused techniques, such as Knowledge Graphs (KG) \cite{Hogan}, enable the deployment of neuro-symbolic reasoning given their capacity for representing knowledge and interactions by means of directed graphs that can represent multiple and heterogeneous relations among entities. In addition, Knowledge Graph Embeddings (KGE) \cite{Martin} is a machine learning task that aims at learning a latent continuous vector space representation (namely, embeddings) of the nodes and edges of a KG, where the nodes represent the road users, the road context, and events, and the edges represent the semantic relations among them. Knowledge completion with KGEs can be used for predicting missing entities (e.g. occluded pedestrians) or relations (e.g. lane change intention) in road scenes that may have been missed by purely data-driven techniques. In this work, we aim at addressing the need to understand and predict road users behaviors by incorporating contextual features into a knowledge-based representation that can also encode other sources of information, such as human knowledge representing driving experience. For that purpose, we propose a neuro-symbolic approach that will combine expressive features (representing road users context) and human experience (in the form of linguistic descriptions and/or rules) in a Knowledge Graph (KG) representation. 
On the one hand, the neural part of this approach will take care of the following tasks: 
i) extracting road users and contextual expressive features using Deep Learning (DL) approaches; 
ii) converting human experience information into entities and relations in the KG using DL generative techniques; 
iii) learn embedded representations for all the entities and relations in the KG using Knowledge Graph Embeddings (KGE). On the other hand, the symbolic part will perform behavior predictions on the grounds of the KG, using Bayesian inference as a downstream task that will enable to perform fully inductive reasoning while providing explainable descriptions of the predicted behaviors. This explainable system has the potential to provide AVs with the capability to best adapt to the driving context, very much in the way human drivers do.  
The rest of this article is organized as follows: \Cref{sec:modelling-road-users-behaviors-using-knowledge-graphs} describes the procedure followed to build a Knowledge Graph that encodes road users’ behavioral models; \Cref{sec:road-users-behavior-prediction} presents the road users’ behavior prediction system using KGE and Bayesian inference; \Cref{sec:explainability} dives into the details of how to achieve explainability with the proposed approach; \Cref{sec:experimantal-results} introduces the implementation and experimental results and \Cref{sec:conclusions-and-future-work} describes the conclusions and future work.

\section{Modelling road user's behaviors using knowledge graphs}
\label{sec:modelling-road-users-behaviors-using-knowledge-graphs}
\noindent 
In the realm of knowledge representation, the \ac{KG} stands as a key tool encoding triples that reveal real-world facts and semantic connections \cite{kgdef}. 
A triple, the fundamental building block within the \ac{KG}, comprises three elements: subject, predicate, and object (alternatively termed as head, relation, and tail). 
Conceptually, a \ac{KG} manifests as a graph wherein edges represent relations and nodes denote entities.
As previously mentioned, one of the applications of \acp{KG} focuses on transforming them into low-dimensional vectors that encode entities and relationships, a technique known as \ac{KGE}. 
The resultant vector is employed for learning and reasoning within embedding-based machine learning models, which can rely on distance-based measures or similarity-based scoring \cite{kgedef}. 
In this work, we employ two different models for \acp{KGE}: a distance-based model known as TransE \cite{transE} and a similarity-based model named ComplEx \cite{complex}.
\renewcommand*{\arraystretch}{1.25}
\begin{table*}[]
\caption{Pedestrian behavior ontology.}
\label{tab:KG_ont_pedestrian}
\begin{center}
\begin{tabular}{|c|>{\centering\arraybackslash}m{6cm}|c|c|c|}
\hline
\textbf{Class} & \textbf{Class Description} & \textbf{Instance} & \textbf{Possible Relation} \\
\hline
Pedestrian &  Generic entity pointing to every child pedestrian  & Pedestrian & Any \\
\hline
Pedestrian ID & Individual Pedestrian ID & Ped1 & HAS\_CHILD \\
\hline
Pedestrian instance ID & ID for a pedestrian at a particular frame & Ped1-30 & INSTANCE\_OF \\
& & & PREVIOUS \\
& & & NEXT \\
\hline
Motion Activity & Pedestrian motion activity & Stand, Walk, Wave, Run, Na & MOTION \\
\hline
Proximity & Pedestrian closeness to the road & NearFromCurb, MiddleDisFromCurb, &\\ & & FarFromCurb & LOCATION \\
\hline
Distance & Pedestrian closeness to the ego-vehicle & TooNearToEgoVeh, NearToEgoVeh, &\\ & & MiddleDisToEgoVeh, FarToEgoVeh & EGO\_DISTANCE \\ & & TooFarToEgoVeh & \\
\hline
Orientation & Pedestrian body orientation  & VehDirection, LeftDirection, & ORIENTATION\\ & & OppositeVehDirection, RigthDirection & \\
\hline
Gaze & Pedestrian attention & Looking, NotLooking & ATTENTION \\
\hline
Cross Action & Crossing behavior of the pedestrian & crossRoad, noCrossRoad & ACTION \\
\hline
\end{tabular}
\label{tab-linguistic}
\end{center}
\end{table*}

In this section, we assess the effectiveness of our approach through two real-world use cases focusing on different road user behaviors, including both pedestrians and drivers. 
Also, an overview of the datasets used in each scenario is provided along with a description of the \acp{KG} creation process.
This includes a description of the associated ontologies, which provide a formal and structured representation of the \ac{KG}, ensuring that it is understandable and explainable. 

\subsection{Pedestrian use case}
\label{sec:ped_use_case}
\noindent
The pedestrian use case focused on predicting whether a pedestrian will cross the road in the next 30 frames. 
The entire pipeline has been trained and tested using two datasets:
\begin{itemize}
    \item \textit{\ac{JAAD}}\footnote{JAAD dataset is publicly available at: \url{https://data.nvision2.eecs.yorku.ca/JAAD_dataset/}}. This dataset comprises 348 short video clips, each extensively annotated to depict various road actors and scenarios across diverse driving locations, traffic, and weather conditions. The dataset annotations encompass spatial, behavioral, contextual, and pedestrian information.
    \item \textit{\ac{PSI}}\footnote{PSI dataset is publicly available at: \url{http://pedestriandataset.situated-intent.net/}}. The dataset comprises 104 training videos, 34 validation videos, and 48 testing videos, collectively covering 196 scenes. It includes bounding box annotations for traffic objects and agents, which are accompanied by text descriptions and reasoning explanations \cite{psi}.
\end{itemize}
The process of modeling pedestrian behavior considered the following characteristics:

\subsubsection{Features extraction and linguistic transformation}
\label{sec:features-extraction-and-linguistic-transformation}
from the mentioned datasets, a set of pedestrian features is extracted using a deep learning approach, and then they are transformed from numerical to linguistic values, as detailed in \cite{kg_ped_predictor}. 
The features extracted for each annotated pedestrian include:

\begin{itemize}
\item \textbf{Motion Activity}: States the motion activity of the pedestrian
 \item \textbf{Proximity to the road}: Transforming from an assessment of road segmentation and pedestrian location to a linguistic representation indicating the pedestrian's proximity to the road in three levels, based on their closeness to it.
\item \textbf{Distance}: Transforming from an estimated distance in meters to a linguistic representation that indicates the pedestrian's proximity to the ego-vehicle.
\item \textbf{Body Orientation}: Transforming from an angle ranging from 0º to 360º to a linguistic representation that encodes the pedestrian's body posture from the perspective of the ego-vehicle.
\item \textbf{Gaze}: Transforming from a binary value to a linguistic indicator that denotes whether the pedestrian is observing the ego-vehicle. 
\end{itemize}

\subsubsection{Pedestrian behavior ontology}
\label{sec:pedestrian-behavior-ontology}
the pedestrian behavior ontology, referred to as PedFeatKG in this study, was built from pedestrian features and was outlined in \Cref{tab:KG_ont_pedestrian}. It encompasses the classes (entities in the KG), their descriptions, the instances of each class representing linguistic values, and the potential relations associated with each class.

\begin{figure*}
     \centering
     \begin{subfigure}[b]{0.48\textwidth}
         \centering
         \includegraphics[width=\columnwidth]{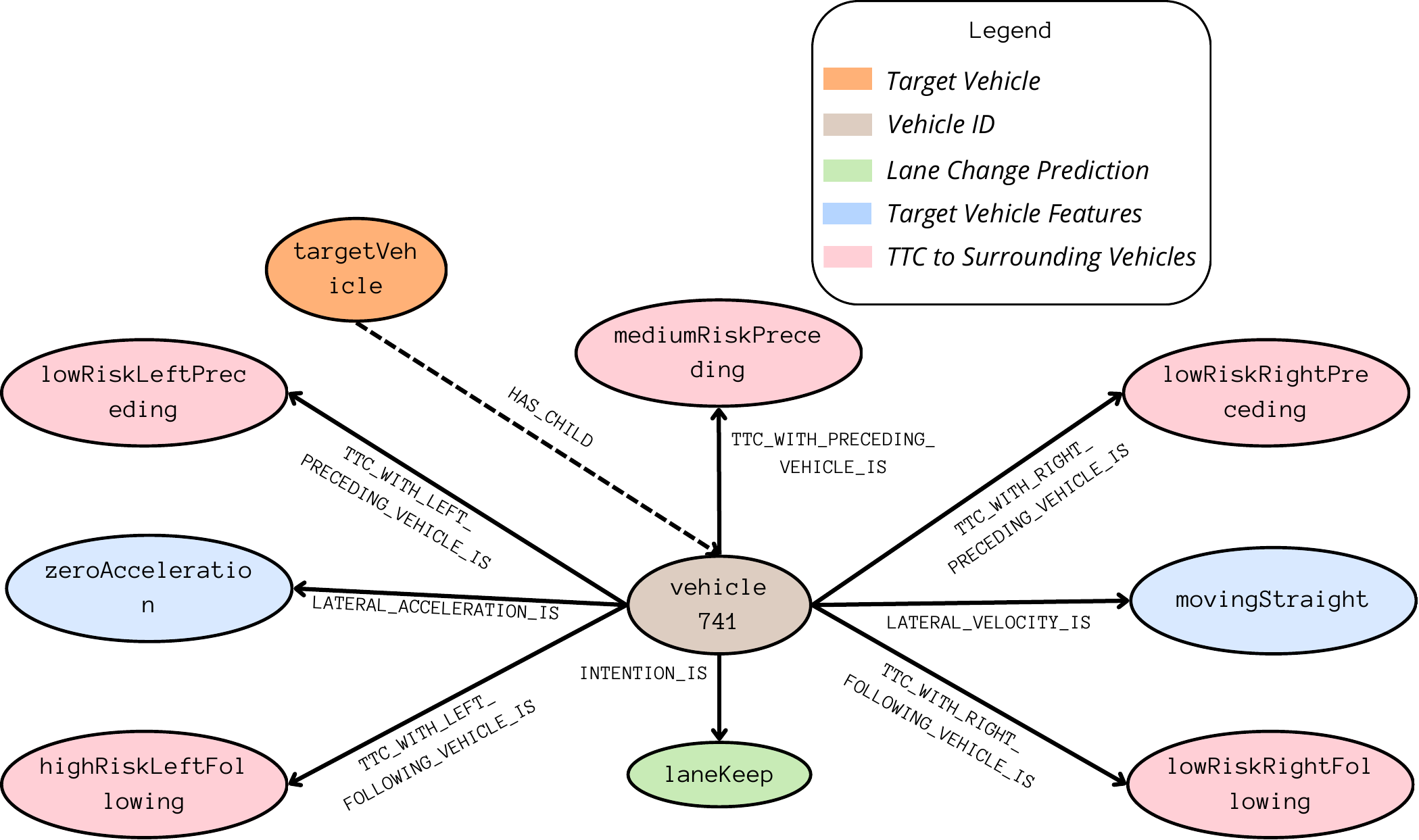}
         \caption{}
         \label{fig:Generated Lane Change KG Instance}
     \end{subfigure}
     \hfill
     \begin{subfigure}[b]{0.48\textwidth}
         \centering
         \includegraphics[width=\columnwidth]{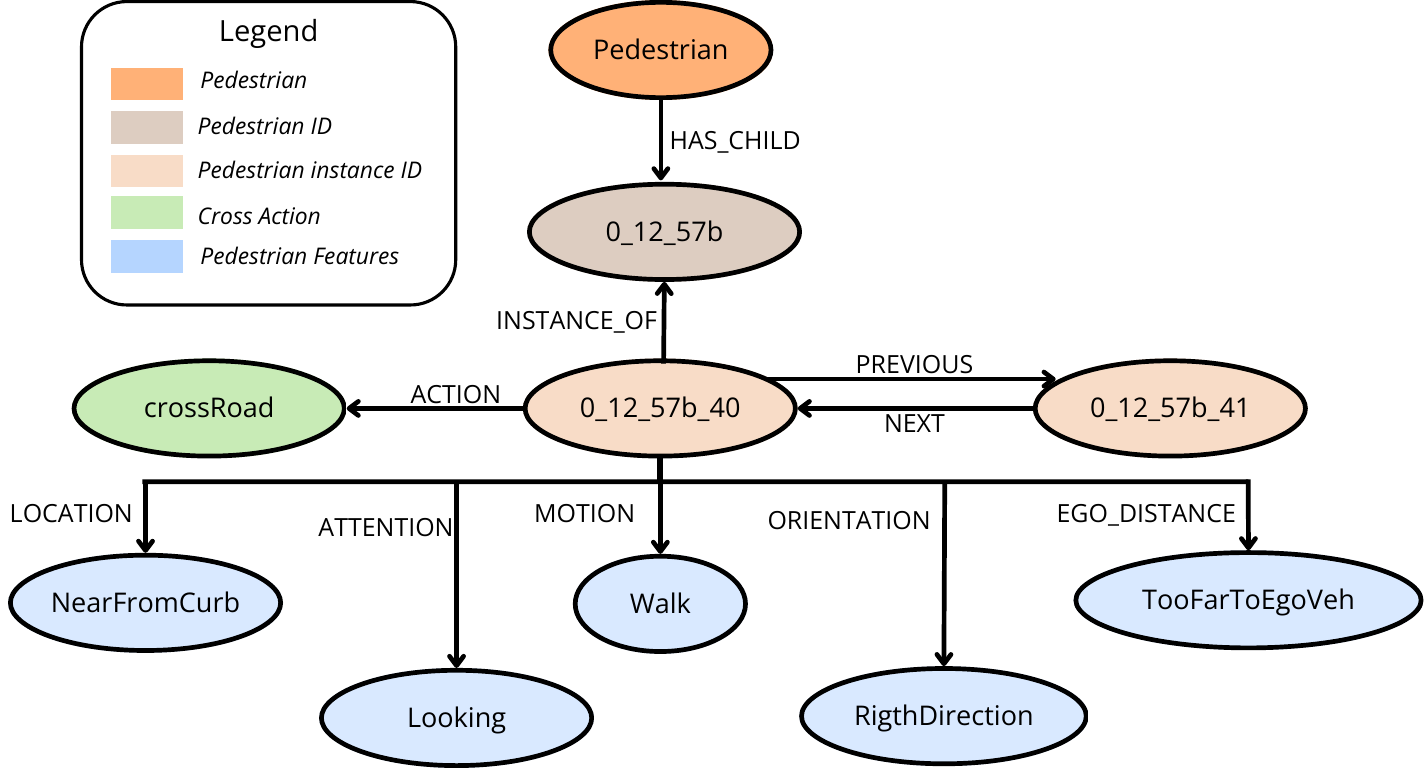}         
         \caption{}
         \label{fig:kg_ped_example}
     \end{subfigure}
     \captionsetup{subrefformat=parens}
     \caption{\subref{fig:Generated Lane Change KG Instance} One KG instance where the vehicle has zero lateral acceleration and has medium \ac{TTC} risk with the preceding vehicle and high \ac{TTC} with the left following vehicle. (b) PedFeatKG from explainable features with 1 instance.}
     \label{fig:twoMainLabel}
\end{figure*}

In the PedFeatKG ontology, each pedestrian from the dataset's training set was represented by a class with a unique ID (noted as \textbf{Pedestrian ID}). At the same time, the Pedestrian ID was associated with a specific pedestrian instance at a particular frame (noted as \textbf{Pedestrian instance ID}). This latter class comprised the pedestrian dataset ID and the frame number. For instance, if a pedestrian has the ID ``ped1'', there will be as many classes as frames considered in the following structure: ``ped1-30'', ``ped1-32'', ``ped1-34'', and so on. Linking the pedestrian ID with the pedestrian instance ID enabled the association of all pedestrian instances, indicating to the \ac{KG} that they represent the same pedestrian across different frames. 
Additionally, each pedestrian instance ID was linked with its previous and next pedestrian instance ID, thus providing temporal association information regarding pedestrian behavior in a road scene within the KG.
Likewise, all pedestrian ID classes extracted were associated with a generalization class called \textbf{Pedestrian}, enabling any specific pedestrian to be linked to a general one. This linkage is considered a path reification link.
On the other hand, each pedestrian instance ID was subsequently linked with the five pedestrian features that represent the pedestrian's state in the following triple format: $<$\textit{pedestrian-instance-ID}, \textit{FEATURE\_RELATION}, \textit{value}$>$. In addition, it can be observed that the pedestrian instance ID was also linked with a crossing behavior, delineated by two possible class values: \textbf{crossRoad} or \textbf{noCrossRoad}. 
\Cref{fig:kg_ped_example} shows a generated KG instance from the PedFeatKG ontology. In this example, it represented the state of the pedestrian with ID \textit{0\_12\_57b} in frame 40, its features, and its future crossing action.
\subsection{Drivers use case}
\label{sec:drivers-use-case}
\noindent
In the lane change prediction use case scenario, the \textit{HighD}\footnote{The dataset is publicly available through the following link \url{https://levelxdata.com/highd-dataset}} dataset \cite{highDdataset} is used. It is a German dataset that was recorded using a camera mounted on a drone, providing a collection of naturalistic top-view scenes of vehicle movements and interactions on German highways. 
The process of modeling lane change behavior considered the following characteristics:

\subsubsection{Linguistic transformation}
the inputs that will be used to construct the KG are vehicle lateral velocity and acceleration, the target vehicle intention, \ac{TTC} with the preceding vehicle, and \ac{TTC} with the left/right preceding and following vehicles. These inputs are extracted from the highD dataset in numerical format. Then, they are converted to linguistic categories. For example, the lateral acceleration numerical value is converted to a category from a set of linguistic categories like accelerating left, zero lateral acceleration, and accelerating right.
To divide each numerical feature into some linguistic categories, some thresholds are determined. 
Following the structure proposed in \cite{manzour2023vehicle}. We used the normal distribution of the lateral velocity and acceleration separation thresholds. 
Regarding the other \ac{TTC} numerical variables, the thresholds are based on the studies in \cite{manzour2023vehicle,saffarzadeh2013general,ramezani2020comparing}.
\subsubsection{Drivers behavior ontology}
\label{sec:pedestrian-behavior-ontology}
the driver behavior ontology, referred to as DriverKG in this study, is based on the reification of nodes and relationships obtained from the HighD dataset, to get reified triples. 
For example, if the vehicle is accelerating to any direction and the \ac{TTC} risk with the left following vehicle is high, then the reified triples will be $<$\textit{vehicle}, \textit{LATERAL\_ACCELERATION\_IS}, \textit{zeroAcceleration}$>$, and $<$\textit{vehicle}, \textit{ TTC\_WITH\_LEFT\_FOLLOWING\_VEHICLE\_IS}, \textit{highRiskLeftFollowing}$>$.

\Cref{tab:KG_LC_classes} shows the KG ontology for the lane change prediction case. The table is divided into four columns. The first column shows the possible classes in the KG. The description of each class is indicated in the second column. The third column shows the possible reified instances that can be assigned to that class, given that the class can take only one instance at a frame. The last column shows the relation that points to that class. In this ontology, a generic entity named \textit{vehicle} is linked to various child vehicles via the \textit{HAS\_CHILD} relation. Each child vehicle is assigned a unique ID (known as \textit{vehicleID}) for each frame. It is important to note that even if it is the same physical vehicle across different frames, it will receive a new \textit{vehicleID} in each frame. Consequently, while both IDs in reality refer to the same vehicle, they are treated as distinct vehicles within the ontology when generating triples and constructing the \ac{KG}.

\renewcommand*{\arraystretch}{1.25}

\begin{table*}[ht]
    \centering
    \caption{Vehicle behavior ontology.}
    \label{tab:KG_LC_classes}
    
    \begin{tabular}{|c|>{\centering\arraybackslash}m{3cm}|c|c|c|}
    \hline    
    \textbf{Class} & \textbf{Class Description} & \textbf{Instance} & \textbf{Possible Relation} \\ \hline
    
    &   & LLC (Left Lane Change)   &\\ 
    intention & Lane changing intention & LK (Lane Keep) & INTENTION\_IS \\ 
    & of the vehicle & RLC ( Right Lane Change)  &\\ \hline
    
    & & movingLeft &\\ 
    latVelocity & Vehicle lateral velocity & movingStraight & LATERAL\_VELOCITY\_IS\\ 
    & & movingRight &\\ \hline
    
    & & leftAcceleration &\\ 
    latAcceleration & Vehicle lateral accelera- & zeroAcceleration (No lateral acceleration) & LATERAL\_ACCELERATION\_IS\\ 
    & tion & rightAcceletion &\\ \hline
    
    & & highRiskPreceding &\\ 
    ttcPreceding & TTC with the preceding  & mediumRiskPreceding &  TTC\_WITH\_PRECEDING\_VEHICLE\_IS \\ 
    & (front) vehicle & lowRiskPreceding &\\ \hline
    
    & & highRiskLeftPreceding &\\ 
    ttcLeftPreceding & TTC with the left & mediumRiskLeftPreceding &  TTC\_WITH\_LEFT\_PRECEDING\_VEHICLE\_IS\\ 
    & preceding (front) vehicle & lowRiskLeftPreceding &\\ \hline
    
    & & highRiskRightPreceding &\\ 
    ttcRightPreceding & TTC with the right & mediumRiskRightPreceding &  TTC\_WITH\_RIGHT\_PRECEDING\_VEHICLE\_IS \\ 
    & preceding (front) vehicle & lowRiskRightPreceding &\\ \hline
    
    & & highRiskLeftFollowing &\\ 
    ttcLeftFollowing & TTC with the left & mediumRiskLeftFollowing &  TTC\_WITH\_LEFT\_FOLLOWING\_VEHICLE\_IS \\ 
    & following (rear) vehicle & lowRiskLeftFollowing &\\ \hline
    
    & & highRiskRightFollowing &\\ 
    ttcRightFollowing & TTC with the right & mediumRiskRightFollowing &  TTC\_WITH\_RIGHT\_FOLLOWING\_VEHICLE\_IS \\ 
    & following (rear) vehicle & lowRiskRightFollowing &\\ \hline
    
    vehicleID & Child vehicle ID which changes every frame & vehicle ID number (e.g. `741') & HAS\_CHILD\\ \hline
    
    vehicle & Generic entity pointing to every child vehicle & -- & Any\\ \hline
    \end{tabular}
\end{table*}

\begin{figure*}[h!]
\centering
\includegraphics[width=0.95\textwidth]{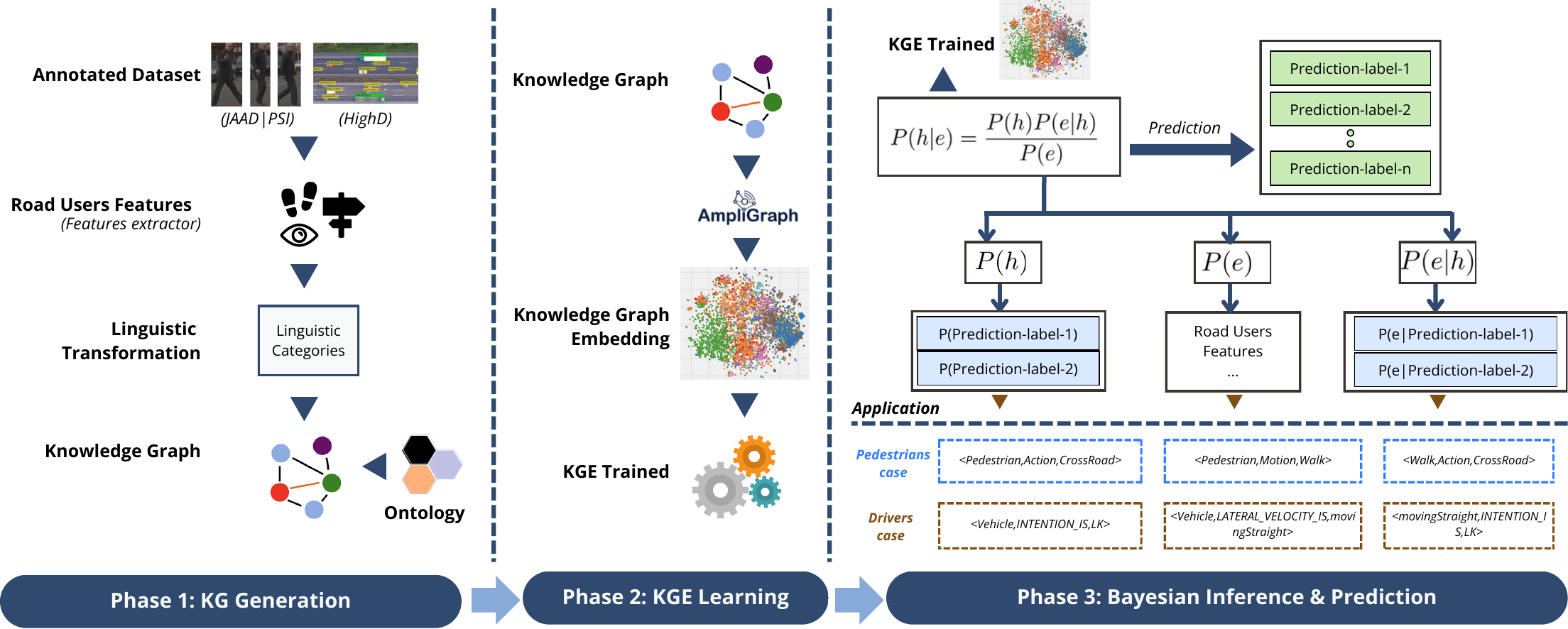}
\caption{Pipeline architecture for modelling road user's behaviors.}
\label{fig:pipeline}
\end{figure*}

\Cref{fig:Generated Lane Change KG Instance} shows a generated KG instance based on the previously mentioned ontology. In this instance, the vehicle with ID 741 is a child of the generic entity vehicle. This can be described in a triple with format $<$\textit{vehicle}, \textit{HAS\_CHILD}, \textit{741}$>$. This child has \textit{latAcceleration} class
assigned to \textit{zeroAcceleration} instance. Also, \textit{741} has \textit{mediumRiskPreceding} and \textit{highRiskLeftFollowing} \ac{TTC}. \textit{vehicle 741} intention is \textit{LK}. 

\section{Road Users' Behavior Prediction Approach}
\label{sec:road-users-behavior-prediction}
\noindent
Both Road Users' Behavior Predictions use cases leverage the proposed architecture based on feature extraction, \acp{KG} and their associated ontologies. 
The overall workflow, depicted in \Cref{fig:pipeline}, comprises three main phases: 1) KG Generation, 2) KGE Learning, and 3) Bayesian Inference and Prediction. This section provides the details on the three phases.

\subsection{Phase 1: KG generation using all types of knowledge}
\label{sec:building-kg-using-all-types-of-knowledge}
\noindent
To capture and encapsulate the data concerning road users' behaviors, the initial step involved extracting the data and features that describe each scene from both the driver's and pedestrian's perspectives. The subsequent step was to convert the extracted features into linguistic values. Following this, utilizing the Ampligraph 2.0.0 library \cite{ampligraph}, the \ac{KG} is constructed in the form of triples, where a set of triples represents the scene in a frame.
Building the KG is a process executed based on a \ac{KG} ontology that generalizes the data applicable to each road scene. This knowledge can originate from various sources and formats, including annotations in datasets, fuzzy rules, and textual explanations concerning road user behavior.

\subsection{Phase 2: KGE Learning}
\noindent
In the second phase, Ampligraph 2.0.0 was employed to construct a \ac{KGE} model using the \ac{KG} generated in the previous phase. 
We used the \textit{ScoringBasedEmbeddingModel} to implement a neural architecture that encodes concepts from a KG into low-dimensional vectors, using a scoring layer such as ComplEx and TransE. 
The training and validation process using the \ac{KGE} model is conducted using the Ampligraph library. 
As a result of this phase, optimal embeddings representing the \ac{KG} are obtained. 
These embeddings are then utilized in the subsequent phase for inference and prediction over the \ac{KG}.

\subsection{Phase 3: Bayesian Inference \& Prediction}
\label{sec:bayesian-inference-and-prediction}
\noindent
Our approach is designed to leverage inductive reasoning by incorporating specific structures, known as reifications, into the knowledge graph by leveraging the properties of the ontology. 
This allows us to perform Bayesian inference (phase 3 in \Cref{fig:pipeline}) on the embeddings derived from the \ac{KG} learning phase (phase 2).
Once these embeddings are obtained, it is then possible to calculate the probabilities of reified triples $P(h,r,t)$ using the evaluation method from the AmpliGraph library.
Then, the Bayes rule in \Cref{eq:bayesian} is used to compute the probability of a hypothesis given some evidence, denoted as $P(h|e)$, where \textit{h} represents the hypothesis (such as the likelihood of a pedestrian intention to cross the road) and \textit{e} stands for the evidence, which in this context is data measured by onboard sensors at a specific moment. 
The datasets (like JAAD for pedestrians and HighD for vehicles) provide this sensory data.
\begin{equation}
\begin{aligned}
    P(h|e)=\frac{P(h)P(e|h)}{P(e)}
\end{aligned}
    \label{eq:bayesian}
\end{equation}
For instance, if the hypothesis is that a pedestrian intends to cross the road and the evidence includes observations such as i) the pedestrian's attention state is looking and ii) the pedestrian's location is near to the vehicle, then the probability of the hypothesis $P(h)$ is determined by evaluating a reified triple, e.g., reifying the intention of a pedestrian to cross the road into the triple $<$\textit{pedestrian, INTENTION\_IS, crossRoad}$>$.
Concerning the calculation of $P(e)$, which actually involves a series of multiple pieces of evidence, we employed equation \Cref{eq:evidence} since each piece of evidence is considered independent. Additionally, each element $e_i$ is reified from the graph.
For example, assuming that the evidence is composed of two elements $e_1$ and $e_2$ indicating respectively that the pedestrian is looking to the ego-vehicle and the pedestrian is near to the ego-vehicle, the associated probabilities are given by the reification of following two triplets $<$\textit{pedestrian, action, looking}$>$ and $<$\textit{pedestrian, egoDistance, nearToEgoVeh}$>$.

\begin{equation}
    P(e) = P(e_1) \times \dots \times P(e_n)
    \label{eq:evidence}
\end{equation}

The probability of the evidence given the hypothesis $P(e|h)$ is calculated based on \Cref{eq:evidence-hypothesis}. 
It is computed as the product of the probabilities of all pieces of evidence assuming the condition that the hypothesis is true. 
These conditioned pieces of evidence are also reified. 
For example, calculating the probability of a pedestrian near to the vehicle given the hypothesis that this pedestrian will cross the road can be reified as $<$\textit{nearToEgoVeh, INTENTION\_IS, crossRoad}$>$. 
The computation of this conditional probability implies that we take for granted that the object entity is a pedestrian who will cross the road. 
Under these conditions, the likelihood of the pedestrian being close to the vehicle in these circumstances will be calculated. 
After that, all computed conditioned probabilities are then multiplied together to determine $P(e|h)$.
\begin{equation}
\begin{aligned}
     P(e|h) = P(e_1,\dots,e_n|h) = P(e_1|h) \times \dots \times P(e_n|h)
\end{aligned}   
    \label{eq:evidence-hypothesis}
\end{equation}
Finally, with all these probabilities available from the graph through the embeddings, the probability of a hypothesis given the evidence $P(h|e)$ can be calculated using the Bayes rule in \Cref{eq:bayesian}.
By analogy, the same concept applies to vehicle lane change prediction. 
For example, let's suppose that we are interested in calculating the probability that a vehicle will keep its current lane given that \ac{TTC} risk with the preceding vehicle is medium and \ac{TTC} risk with the left following vehicle is high. 
Then, the probability $P(h)$ is computed by evaluating the triple $<$\textit{vehicle, INTENTION\_IS, LK}$>$ from the \ac{KGE}.
In a similar way, the probability $P(e|h)$ is computed by the multiplication of the two evaluated triplets (1) $<$\textit{mediumRiskPreceding, INTENTION\_IS, LK}$>$ and (2) $<$\textit{highRiskLeftFollowing, INTENTION\_IS, LK}$>$.
Finally, $P(h|e)$ is calculated using \Cref{eq:bayesian}. This process combines structured knowledge representation with probabilistic Bayesian inference to make predictions based on observed data.
This process is repeated for each label, so the probability of \textit{LLC} is computed given the generated linguistic inputs, the same computation is done for \textit{LK} and \textit{RLC}, and the score with the highest probability will be the model's prediction. In the pedestrian use case, the probabilities of \textit{crossRoad} and \textit{noCrossRoad} are computed given the generated linguistic inputs, and the highest probability will be considered as the model's prediction.
 
\section{Explainability}
\label{sec:explainability}
\noindent 
Although the \ac{KG} inherently offers some insights into explainability, in our study we chose to integrate additional tools to support prediction explanations. 
Hence, we incorporated the fuzzy logic approach and the \ac{RAG} technique. 
In this section, we provide a detailed overview of both approaches to enhance explainability.

\subsection{Explanation using fuzzy rules}
\label{sec:explanation-using-fuzzy-rules}
\begin{figure}[t]
    \centering
    \includegraphics[width=\columnwidth]{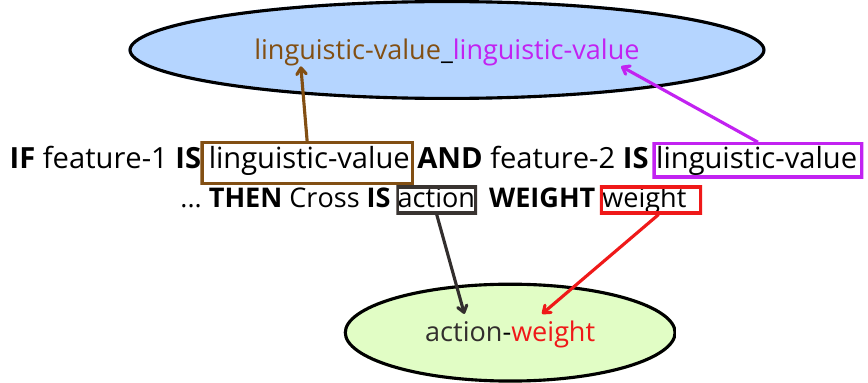}
    \caption{Fuzzy rule conversion definition.}
    \label{fig:fuzzy_to_kg}
\end{figure}

\begin{figure}[b]
    \centering
    \includegraphics[width=\columnwidth]{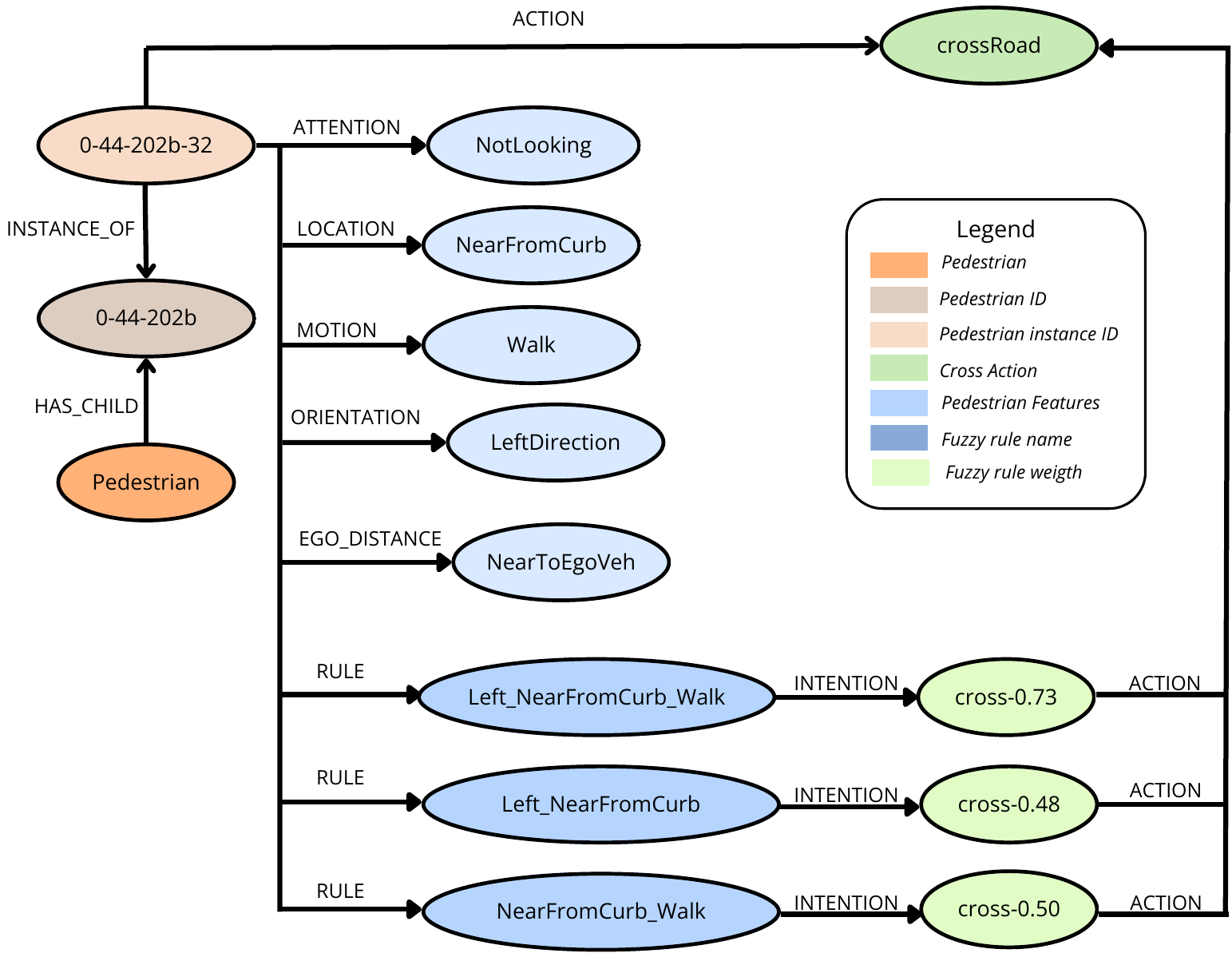}
    \caption{PedFeatRulesKG from explainable features with 1 instance.}
    \label{fig:pedfeatrules_instance}
\end{figure}

\noindent
To integrate explainability into our approach we first adopted a fuzzy logic approach, characterized by its multi-valued nature that mirrors human thought and interpretation.
The generation of fuzzy rules entailed a rule mining process employing various learning algorithms and classification systems. 
In this study, we employed the IVTURS-FARC system \cite{ivturs}, which utilizes interval-valued restricted equivalence functions to enhance rule relevance during inference. 
The learning process of this system involved the FARC-HD algorithm \cite{farchd} for fuzzy association rule extraction. 
The rule mining process began with the extracted pedestrian features from \ac{JAAD} and \ac{PSI} datasets, then, employing the mentioned methods, we extract a set of rules structured as follows:

\begin{equation}
    \begin{split}
        Rule \; R_{j} : & \quad if \;x_{1}\; is \;A_{j1}\; and ... and\;  x_{n}\; is\; A_{jn} \\
                        & \quad then\; Class\; = C_{j}\; with\; RW_{j}
    \end{split}
\end{equation}
In the given expression \(R_{j}\) denotes the label of the  \(j\)th rule, \(x = (x_{1},...,x_{n})\) represents a n-dimensional pattern vector (related to pedestrian features in our context), \(A_{ji}\) stands for an antecedent fuzzy set indicating a linguistic term, \(C_{j}\) signifies the class label, and \(RW_{j}\) denotes the weight of the rule \cite{rulesdef}. 
In this work, we initially applied the fuzzy logic approach to the pedestrian use case. 
In the pedestrian use case the rule mining process produced 60 rules from the PSI data, while 51 rules originated from the JAAD data. To gain insights into pedestrian behavior, we utilized these rules and integrated them into the \ac{KG}. 
Each fuzzy rule was transformed into two classes (as illustrated in  \Cref{fig:fuzzy_to_kg}) which were focused on: 1) combining all feature values into one class and 2) integrating the crossing action and rule weight. 
Introducing these two new entities in the KG required to modify the associated ontology to reflect the changes. Therefore, we established another ontology labeled as \textbf{PedFeatRulesKG}, where pedestrians, alongside their feature states, are linked with rules relevant to their states, thus offering more insights into their behavior. 
The PedFeatRulesKG ontology can be considered as an extension of the PedFeatKG ontology (see \Cref{sec:pedestrian-behavior-ontology}). 
\Cref{fig:pedfeatrules_instance} illustrates an example of the mentioned ontology for a pedestrian in a frame, showcasing the explainable features associated with its state and the correlation with fuzzy rules that apply to this state.

\subsection{Explanation using RAG}
\label{sec:explanation-using-rag}
\begin{figure}[t]
    \centering
    \includegraphics[width=\columnwidth]{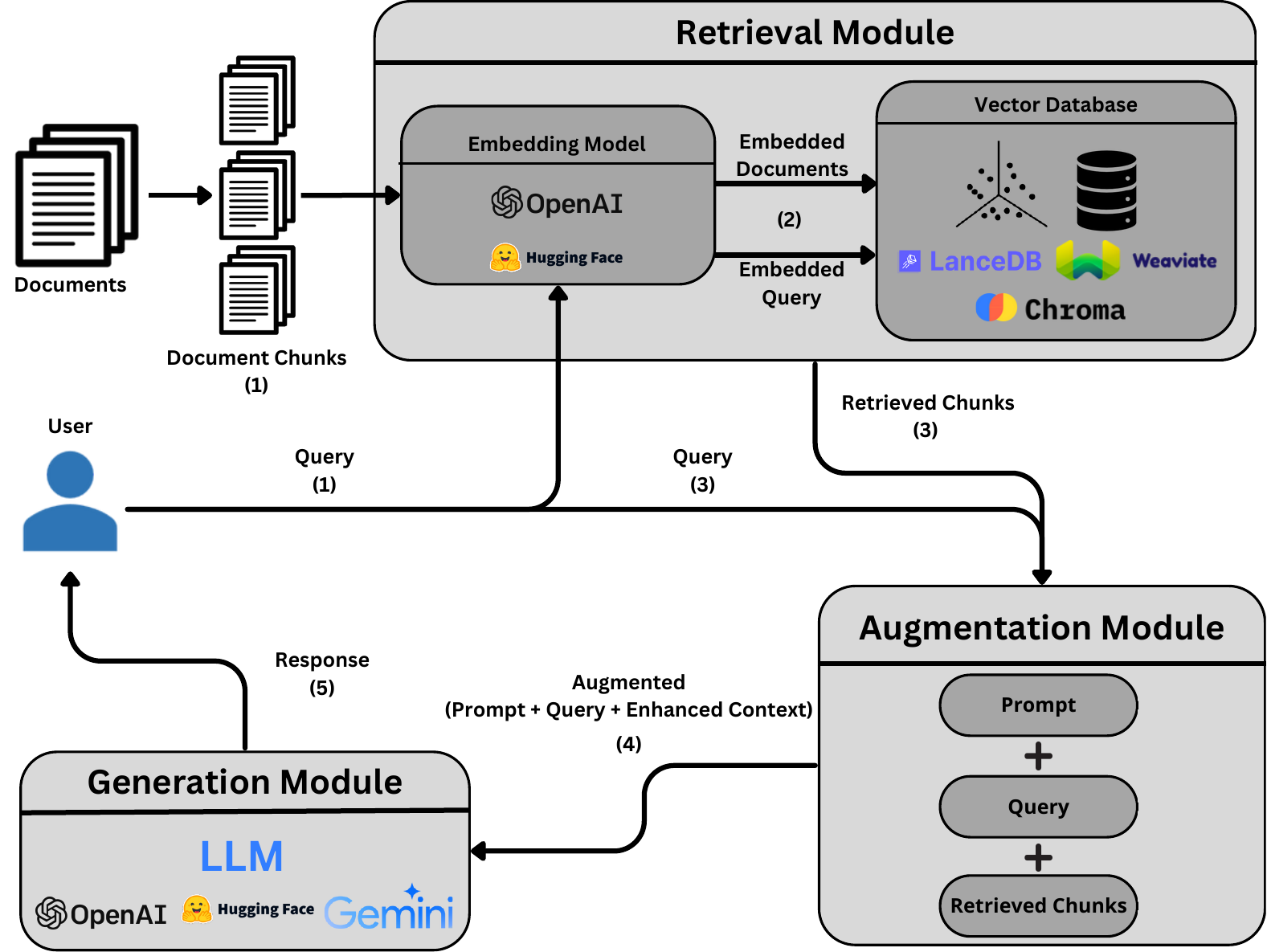}
    \caption{RAG workflow (the numbers show the arrangement of the RAG process flow throughout the figure).}
    \label{fig:RAG}
\end{figure}

\noindent
\acf{RAG} is a machine learning model that combines the power of pre-trained language models with the ability of a retrieval system to retrieve relevant information from a large database and then use this information to generate precise and contextually rich responses to the user's query \cite{lewis2020retrieval}.
This large database can be a public database or a private database that is specified for a certain domain. 
The private database can be obtained by leveraging external knowledge sources, which enhances the model's responses to be more accurate, informative, and specific to that certain task or domain.
The \ac{RAG} process, schematically depicted in \Cref{fig:RAG}, consists of three modules: retrieval module, augmentation module, and generation module.
First, the document which forms the source database is divided into chunks.
These chunks, transformed into vectors using an embedding model like OpenAI or open source models available from Hugging Face community, are then embedded into a high-dimensional vector database (e.g., Chroma and LlamaIndex).
When the user inputs a query, the query is embedded into a vector using the same embedding model. Then, chunks whose vectors are closest to the query vector, based on some similarity metrics (e.g., cosine similarity) are retrieved. This process is contained in the retrieval module shown in the figure.
After that, the retrieved chunks are augmented to the user's query and the system prompt in the augmentation module. This step is critical for making sure that the records from the retrieved documents are effectively incorporated with the query.
Then, the output from the augmentation module is fed to the generation module which is responsible for generating an accurate answer to the query by utilizing the retrieved chunks and the prompt through an LLM (like chatGPT by OpenAI, hugging face, and Gemini by Google).
We used the RAG procedure described in \Cref{fig:RAG} for both our use cases, pedestrian crossing behavior prediction and vehicle lane change prediction. The only difference is the source of knowledge and the query that will be fed to the retrieval module.

\section{Implementation and experimental results}
\label{sec:experimantal-results}
\noindent
In this section, we provide a detailed description of how our behavior prediction approach was implemented for both pedestrian and driver scenarios. 
While both scenarios share a common \ac{KGE} learning strategy and the Bayesian inference scheme, as described in previous sections, here we highlight specific details associated with each scenario, beginning with the datasets and experiments considered.
Subsequently, we present and analyze the extensive experimental activity we performed to evaluate the results with respect to other state-of-the-art approaches. 

\subsection{Implementation details - Pedestrian use case}
\label{sec:imp-pedestrian-use-case}
\noindent
The implementation of our approach for the pedestrian use case can be delineated in three segments: 1) Feature extraction, 2) Modeling pedestrian behaviors utilizing KG and 3) Explainability using RAG. 
Firstly, we build a modular architecture employing Python and PyTorch to extract the features specified in \Cref{sec:ped_use_case}. 
These extractions were facilitated by various neural networks and processed estimations as it is briefly described in  \Cref{table:ped_feat_extraction}.
All implementation details are expounded upon in \cite{melo2023experimental}. 
\begin{table}[h!]
    \centering
    \caption{Pedestrian Features Extraction.}
    \label{table:ped_feat_extraction}
    \begin{tabular}{|p{1.5cm} |p{1.5cm}|p{4cm}|}
    \cline{1-3} 
    \textbf{Feature} & \textbf{Extraction type} & \textbf{Description}  \\ \hline
Motion activity & Neural network & We implemented a transformer architecture that processes the 2D body pose and outputs the pedestrian action. \\ \hline
    Proximity to the road & Neural network and estimation  & From YOLOPv2\cite{yolopv2} was obtained the drivable road area segmentation and lane detection. Based on an experimental minimum distance it is estimated whether the pedestrian is near to the road or not.\\ \hline
    Distance & Estimation & Estimated using the triangle similarity \\ \hline
    Orientation & Neural network &  Using the PedRecNet\cite{pedrecnet} the joint positions of the human body and the body orientation from the azimuthal angle  \( \varphi \) were obtained.  \\ \hline
    Gaze & Estimation &  We used the 2D body pose detection and the positions of the nose, left eye, and right eye keypoints.\\ \hline    
    \end{tabular}
\end{table}

\begin{figure*}[ht]
    \centering
    \includegraphics[width=\linewidth]{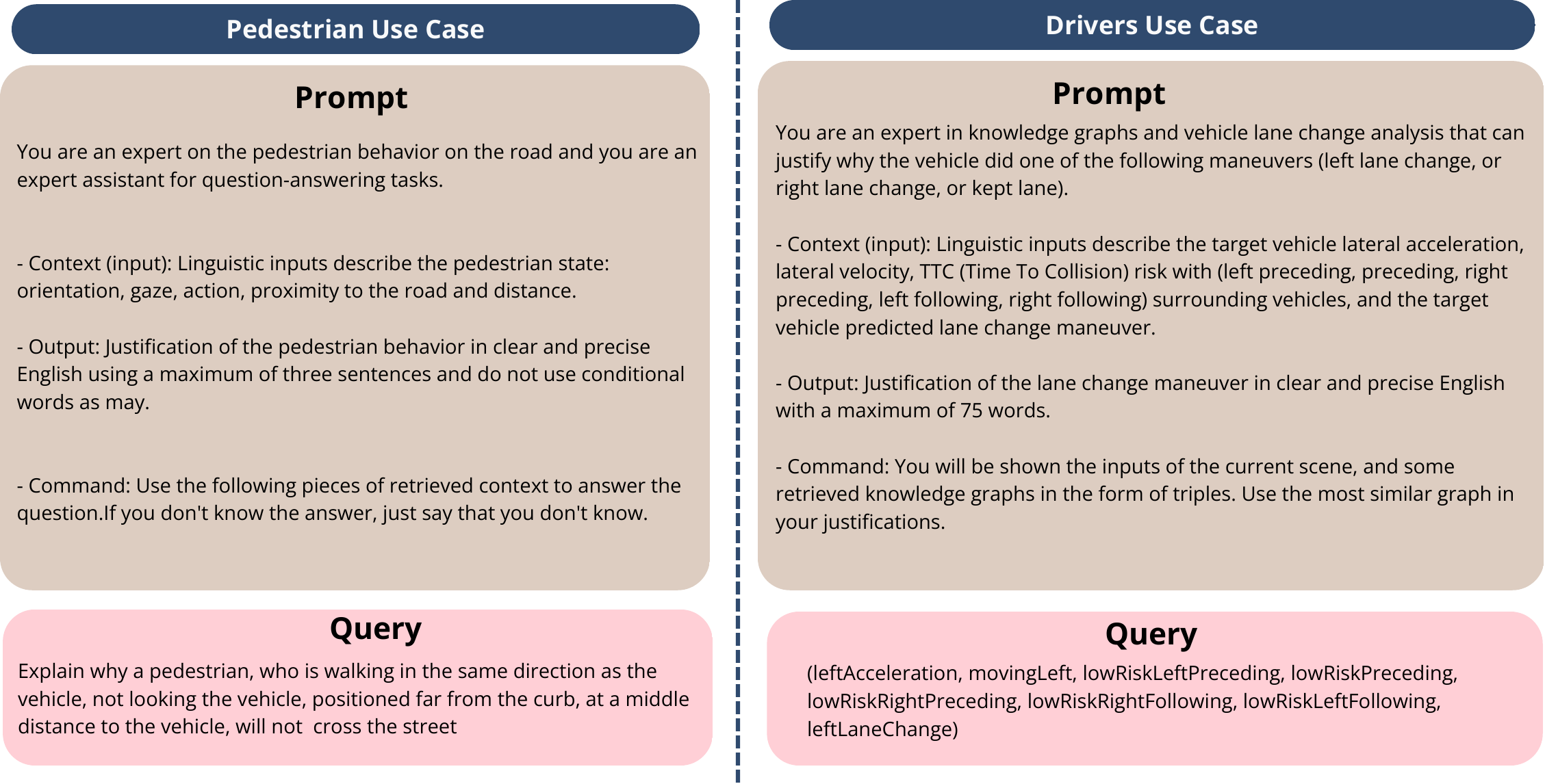}
    \caption{RAG prompt example for road users behavior.}
    \label{fig:rag_prompt}
\end{figure*}

\noindent
Secondly, we employed Python, TensorFlow, and the AmpliGraph 2.0.0 library to execute the pipeline for modeling pedestrian behaviors utilizing \ac{KG}. 
In this implementation, we utilized the ComplEx scoring model, the Adam optimizer, and the SelfAdversarialLoss.
In terms of training parameters, we utilized an embedding size of \(k = 150 \). The number of corruptions generated during training ranges from 5 to 20, depending on the quantity of triples and the dataset. 
Additionally, we set \(learningRate = 0.0005 \), \(batchSize = 10\,000 \), and implemented an early stopping criterion using the \ac{MRR} as is described in \cite{kg_ped_predictor}.
The pedestrian use case involved a set of videos for both training and testing purposes. 
In the case of \ac{JAAD}, 136 videos were utilized for training and 35 for testing, whereas for \ac{PSI}, 104 videos were used for training and 48 for testing. 
From these datasets and the ontologies described before (PedFeatKG and PedFeatRulesKG), two \acp{KG} were generated, each composed of a specific number of triples and entities, as detailed in \Cref{table:triples_ped_case}.
\begin{table}[]
    \centering
    \caption{Number of triples in the experimental setup.}
    \label{table:triples_ped_case}
    \begin{tabular}{|c|l|l|}
    \cline{1-3} 
    \multicolumn{1}{|l|}{\textbf{Dataset}} & \multicolumn{1}{c|}{\textbf{Ontology}} & \multicolumn{1}{c|}{\textbf{Triples}} \\ \hline
    \multirow{2}{*}{PSI}        &  PedFeatKG        &  $238\,795$  \\ 
                                &  PedFeatRulesKG       &  $302\,574$  \\ \hline
    \multirow{2}{*}{JAAD}       &  PedFeatKG        & $139\,624$  \\ 
                                &  PedFeatRulesKG       &  $197\,381$   \\ 
    \hline
    \end{tabular}
\end{table}
The performance evaluation of the proposed pipeline was conducted using precision, recall, and F1-score metrics. Precision was calculated as the ratio of correct positive predictions to the total predicted positives. Recall represents the ratio of correct positive predictions to the total positive examples, while the F1-Score is the harmonic mean of precision and recall.
Finally, the third segment was focused on RAG implementation. To accomplish this explainability module, we utilized pedestrian features extracted from the JAAD and PSI datasets to generate a human-readable document that incorporates a basic explanation of why pedestrians have or do not have the intention to cross the road, for instance: \textit{"The pedestrian \underline{will not cross} the street \underline{because}, the pedestrian is looking, is oriented to the left, is running, is at a moderate distance from the road and the vehicle is too far"}.
Then, we utilized this document containing all descriptions in the RAG module, based on the LangChain framework. Within this module, we segmented the document into chunks and transformed them into embeddings using the OpenAI model. The embeddings were then stored in the Chroma vector database. Subsequently, the final response was generated using the OPENAI GPT-4 \ac{LLM}, based on a prompt tailored for the pedestrian use case and a query derived from the pedestrian features within the prediction frame, as detailed in  \Cref{fig:rag_prompt}.

\subsection{Implementation details - Drivers use case}
\label{sec:imp-drivers-use-case}
\noindent
This section addresses two main topics regarding the driver use case. The first one focuses on training the generated \ac{KG} obtained from phase 1. The training utilizes the Ampligraph library, which creates the \ac{KGE} from the HighD dataset. The second topic discusses the generation of explanations using the \ac{RAG} technique.
The dataset was divided based on tracks, to ensure a clear distinction between training, validation and testing data, taking into account the vehicles' behaviors across different tracks. 
This division was crucial to prevent the overlap of behaviors, especially since vehicles on the same track could exhibit similar behaviors, such as vehicles moving to the right because there is an exit on the right at the end of the road.
Consequently, the dataset was organized so that the first 48 tracks (80\% of the data), were allocated for the training and validation phases. 
The remaining 12 tracks (20\% of the data) were reserved exclusively for testing.
A variety of triple counts are explored for validation, including $500$, $1000$, $2000$, $4000$, and $10\,000$ triples. 
Despite this range, the evaluation score during testing remained consistent across these different triple counts. 
Therefore, a decision was made to proceed with $2000$ triples for validation, leveraging the \textit{train\_test\_split\_no\_unseen} function provided by the Ampligraph library to facilitate this choice. 
The final distribution of triples for the dataset was established as $351\,736$ for training, $2000$ for validation, and $12\,222$ for testing.
Two distinct scoring models are compared: \textit{TransE} and \textit{ComplEx}. To ensure a fair comparison, the training parameters are fixed. 
This includes setting the embedding size (k) to $100$, employing the Adam optimizer with learning rate = $0.0005$, utilizing the SelfAdversarialLoss, generating five negative triples for every positive triple by corrupting both the subject and object and setting a batch size of $10\,000$. 
Additionally, validation parameters were specified, with a burn-in period and frequency both set at five, alongside a validation batch size of $100$. 
An early stopping criterion is also used to monitor the \ac{MRR} metric during validation, with a patience threshold of five validation epochs. F1-score is the used evaluation metric to choose the best model. Also, precision and recall metrics are used for comparison purposes results with other works.
Regarding the \ac{RAG} section the data is divided into chunks with size of 384 tokens.  Each chunk represents a \ac{KG} of one sample in the form of triples. Chunks were transformed into embedding vectors using \textit{all-MiniLM-L6-v2} Hugging Face embedding model and stored in Chroma vector database. 
After that, OpenAI GPT-4 \ac{LLM} was used in the generation module to generate the final response.
The query is formed by extending the linguistic inputs which are fed to the \ac{KGE} and Bayesian inference model with the lane change prediction output obtained from that model. Then, this query is fed to the \ac{RAG} model. 
Figure \Cref{fig:rag_prompt} shows the used system prompt with an example where the query is provided in order to guide the \ac{LLM} model when generating the responses.
\subsection{KG-based Prediction Results - Pedestrian use case}
\label{sec:res-pedestrian-use-case}
\noindent
\begin{table}[b]
    \centering
    \caption{Comparing the pedestrian behavior predictor with various methods (The table includes the available results).}
    \label{table:ped_results}
    \subcaption{ \(JAAD_{test} \)}    
    \begin{tabular}{lcccc}
        \textbf{Model} & \textbf{F1}  & \textbf{Precision} & \textbf{Recall} & \textbf{Accuracy} \\
        \hline
        \acs{C3D} & 0.65 & 0.57 & 0.75 & 0.84 \\
        \acs{PCPA} & 0.68 & - & - & \textbf{0.85} \\
        Decision Tree & 0.78 & 0.78 & 0.78 & 0.78\\
        Fuzzy Logic & 0.75 & 0.69 & 0.81 & 0.69 \\
        PedFeatKG & 0.86 & 0.77 &  \textbf{0.96} & 0.79 \\
        PedFeatRulesKG & \textbf{0.87} & \textbf{0.86} & 0.88 & 0.83\\
        \hline
    \end{tabular}

    \bigskip

    \subcaption{ \(PSI_{test} \)}        
    \begin{tabular}{lcccc}
        \textbf{Model} & \textbf{F1}  & \textbf{Precision} & \textbf{Recall} & \textbf{Accuracy} \\
        \hline
        \acs{eP2P} & 0.66 & - & - & 0.76 \\
        Ours Black Box & 0.75 & 0.74 & 0.75 & 0.62 \\
        Decision Tree & 0.63 & 0.63 & 0.63 & 0.63 \\
        Fuzzy Logic & 0.72 & 0.74 & 0.70 & 0.59 \\
        PedFeatKG & 0.81 & \textbf{0.75} & 0.89 & 0.69\\
        PedFeatRulesKG & \textbf{0.84} &   \textbf{0.75} & \textbf{0.94} & \textbf{0.72} \\
        \hline
    \end{tabular}
\end{table}
The performance of our knowledge-based predictor approach was evaluated over the test set of each dataset. The results provided were compared with the following methods:
\begin{itemize}
    \item In \ac{JAAD}: 
    \begin{itemize}
        \item \ac{C3D}: it is a state of the art model that utilizes  RGB frames and a \ac{fc} layer to generate the final prediction \cite{cd3} \cite{benchcrossing}.  
        \item \ac{PCPA}: it is a state of the art model that integrates a 3D convolutional branch for encoding visual information, alongside individual RNNs to process various features \cite{benchcrossing}.
        \item Decision Tree: to evaluate this technique, we used the simple implementation of Decision Trees provided by the KNIME Anaylitics Platform, using the Gini Index as quality meter, a minimum records per node set as 4 and with Minimal Description Length (MDL) pruning method activated.
        \item Fuzzy logic: as mentioned in Section \Cref{sec:explanation-using-fuzzy-rules}, we utilized the IVTURS-FARC method to extract a set of fuzzy rules and membership functions. Subsequently, employing a \ac{TS} inference system implemented in Python, we generated predictions.
    \end{itemize}
    \item In \ac{PSI}:
    \begin{itemize}
        \item \ac{eP2P}: it is a state-of-the-art model that leverages context features and LSTM encoder-decoder modules to forecast pedestrian intentions and trajectories \cite{psi}.
        \item Ours Black Box: we developed the black box to participate in the IEEE ITSS Student Competition on Pedestrian Behavior Prediction, which took place in 2023 at ITSC2023. This black box employed transformer encoding blocks, pedestrian features, and a many-to-one attention layer to predict crossing intention.
        \item Decision Tree: we used the process described above to generate predictions through this approach.
        \item Fuzzy Logic: we used the process described above to generate predictions through fuzzy logic approach.
    \end{itemize}
\end{itemize}
According to the results presented in \Cref{table:ped_results}, experiments conducted on both datasets, PSI and JAAD, demonstrate that both KG models outperform other methods focusing on ``black box'' strategies or explainability. Specifically, in the case of JAAD, the KG models significantly enhance performance in terms of F1-score, with PedFeatRulesKG showing a 22\% improvement compared to C3D and a 19\% improvement compared to PCPA. Similarly, compared to the decision tree and fuzzy logic approach, our method demonstrates improvements of 9\% and 12\%, respectively. While improvements in precision and recall are also evident in our approach, accuracy values in ``black box'' methods are higher.
In the case of PSI, improvements are evident in terms of F1-scores, precision, and recall, while accuracy remains higher with the ``black box'' approach. Specifically, PedFeatRuleKG shows improvements in the F1-score compared to eP2P, our black box method, the decision tree, and the fuzzy logic approach, by 18\%, 9\%, 21\%, and 12\%, respectively.
In addition, in both datasets, both KG models yield similar results. 
However, the best performance is achieved by the \ac{KG} incorporating pedestrian features and fuzzy rules (PedFeatRulesKG). 
This highlights the importance of integrating various sources of information into the \ac{KG} to enhance pedestrian behavior predictions. Moreover, the inclusion of fuzzy rules enhances the robustness of the \ac{KG} and provides additional evidence, which is included in the Bayesian inference process, offering clues that differentiate between crossing and non-crossing predictions.
The presented results demonstrated that our approach, in addition to offering a novel strategy for incorporating explainability into pedestrian behavior prediction tasks, included a pipeline based on \ac{KG} and Bayesian inference that delivered outstanding performance. Furthermore, it is important to highlight that the incorporation of fuzzy rules to enrich the \ac{KG} served as a valuable complement to Bayesian inference. 
\begin{figure*}[]
    \centering
    \includegraphics[width=\linewidth]{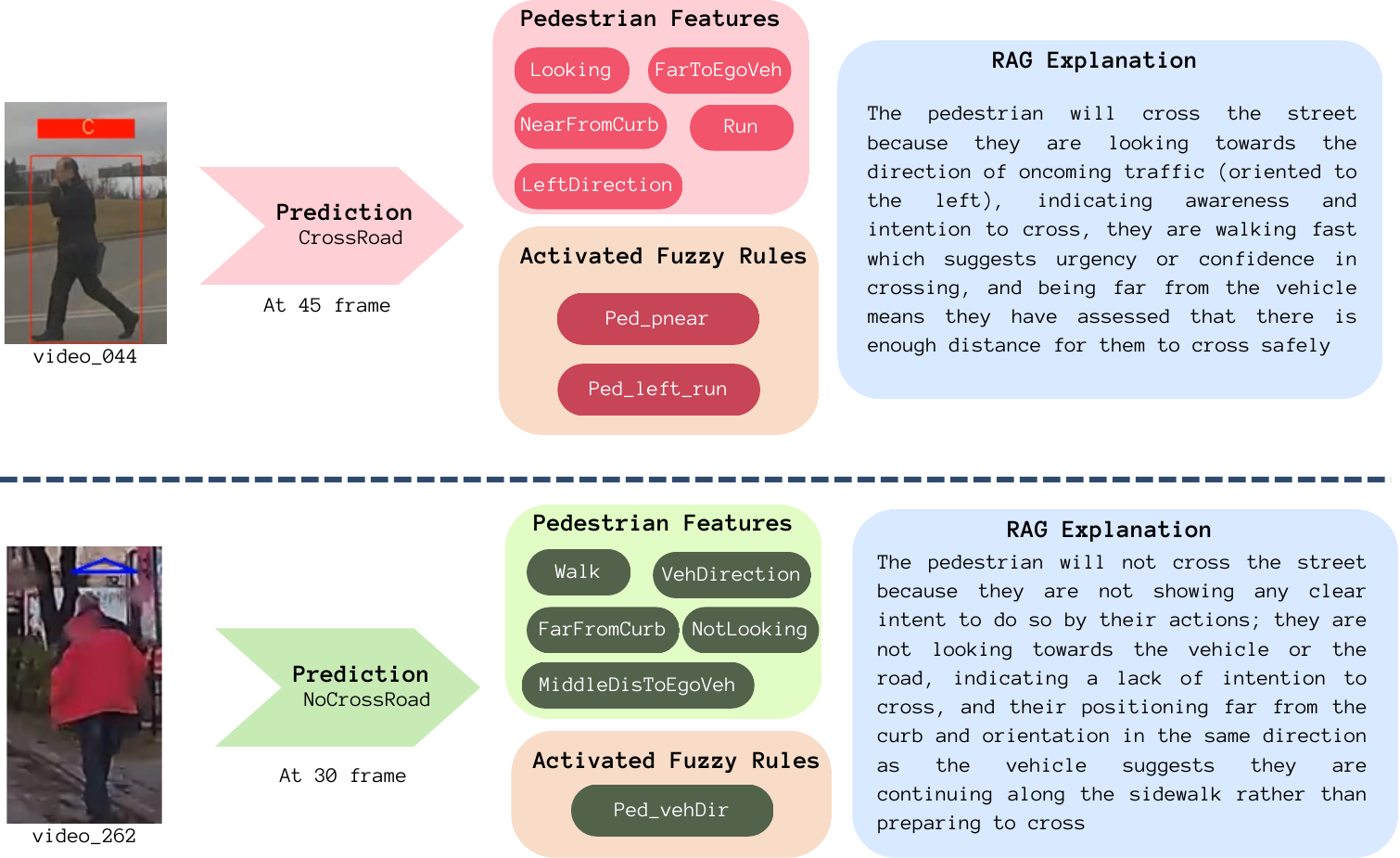}
    \caption{Examples of prediction explainability from JAAD dataset.}
    \label{fig:explain_ped}
\end{figure*}
\subsection{KG-based Prediction Results - Drivers use case}
\noindent
After incorporating the \ac{KGE} embedding and applying Bayesian inference, we tested the model against the last 12 tracks of the HighD dataset. 
Initial tests compared the F1-scores of the TransE and ComplEx models three seconds before a lane change. 
With the TransE model achieving an f1-score of 93.60\% and the ComplEx model only reaching 12\%, the TransE model demonstrated superior performance. 
Thus, subsequent experiments and discussions are focused on the TransE model.
We conducted tests at varying intervals before a lane change, ranging from 1 to 4 seconds, in one-second increments. 
The results are stated in \Cref{tab:results}, showing that the model's F1-score remains above 90\% for three seconds before the lane change event. 
We compared our model's F1-score with those reported in existing literature using the HighD dataset, as detailed in \Cref{tab:comparison}. 
This comparison reveals that our model performance is close to the scores in studies \cite{xue2022integrated} and \cite{gao2023dual} within the initial 0.5 to 1-second time frame, showing only a 1\% margin difference. 
However, starting from the 1.5-second and higher, our model consistently outperforms the aforementioned studies, maintaining an F1-score above 97\% at 2.5 seconds, above 90\% for up to three seconds, and over 80\% for 3.5 seconds before crossing the lane marking.
To further evaluate the effectiveness of our proposal, we compared it to a standard machine-learning technique using a simple Decision Tree implementation in the KNIME Analytics platform.
Our results showed that both approaches produced similar results in the range from 1 second to 2 seconds, indicating that our proposal can achieve comparable performance to traditional machine learning techniques in certain scenarios.
However, our proposal began to outperform the Decision Tree implementation at 3.0 seconds before crossing, highlighting the superior predictive capabilities of our Bayesian inference schema. 
This result demonstrates the potential of our approach to provide more accurate and reliable predictions in complex scenarios, where traditional machine learning techniques may struggle to capture the nuances and uncertainties of human behavior.

\begin{table}[]
\centering
\caption{Precision, recall, and F1-score metrics of the predictions obtained from our proposed model at different instants.}
\label{tab:results}
\resizebox{\columnwidth}{!}{%
\begin{tabular}{cccc|c}
\multicolumn{4}{c|}{\textbf{Our Proposal}}                                          & \textbf{Decision Trees}                    \\ \hline
1 Second  & \textbf{Precision (\%)} & \textbf{Recall (\%)} & \textbf{F1-score (\%)} & \multicolumn{1}{l}{\textbf{F1-score (\%)}} \\ \hline
LK        & 98.33                   & 96.96                & 97.64                  & 97.69                                      \\
LLC       & 97.98                   & 97.50                & 97.74                  & 98.03                                      \\
RLC       & 97.00                   & 99.42                & 98.19                  & 97.91                                      \\ \hline
Macro avg & 97.77                   & 97.96                & \textbf{97.86}                  & \underline{\textbf{98.88}}                                  \\ \hline
2 Seconds & \textbf{Precision (\%)} & \textbf{Recall (\%)} & \textbf{F1-score (\%)} & \multicolumn{1}{l}{\textbf{F1-score (\%)}} \\ \hline
LK        & 98.86                   & 96.96                & 97.95                  & 97.19                                      \\
LLC       & 97.50                   & 99.15                & 98.32                  & 97.36                                      \\
RLC       & 96.52                   & 98.66                & 97.58                  & 97.04                                      \\ \hline
Macro avg & 97.66                   & 98.25                & \underline{\textbf{97.95}}                  & \textbf{97.20}                                      \\ \hline
3 Seconds & \textbf{Precision (\%)} & \textbf{Recall (\%)} & \textbf{F1-score (\%)} & \multicolumn{1}{l}{\textbf{F1-score (\%)}} \\ \hline
LK        & 92.53                   & 96.96                & 94.70                  & 93.19                                      \\
LLC       & 95.71                   & 91.77                & 93.70                  & 90.38                                      \\
RLC       & 95.46                   & 89.50                & 92.38                  & 91.71                                      \\ \hline
Macro avg & 94.56                   & 92.74                & \underline{\textbf{93.60}}                  & \textbf{91.76}               \\ \hline
4 Seconds & \textbf{Precision (\%)} & \textbf{Recall (\%)} & \textbf{F1-score (\%)} & \multicolumn{1}{l}{\textbf{F1-score (\%)}} \\ \hline
LK        & 69.63                   & 96.96                & 81.05                  & 79.04                                      \\
LLC       & 91.30                   & 46.00                & 61.16                  & 46.19                                      \\
RLC       & 88.75                   & 42.39                & 57.37                  & 53.61                                      \\ \hline
Macro avg & 83.22                   & 61.78                & \underline{\textbf{66.52}}                  & \textbf{59.61}                                     
\end{tabular}%
}
\end{table}

\begin{table}[]
\renewcommand{\arraystretch}{1.2}
\caption{Comparison with other models using the F1-score (\%) metric.}
\begin{center}
\begin{tabular}{|c|c|c|c|c|c|}
\hline
\backslashbox{Algorithm}{Prediction\\Time} & 0.5s & 1.0s & 1.5s & 2.0s\\
\hline
\cite{xue2022integrated}   & 98.20  & 97.10  & 96.61 & 95.19\\
\hline
\cite{gao2023dual}   & \underline{\textbf{99.18}}  & \underline{\textbf{98.98}}  & 97.56 & 91.76 \\
\hline
Ours  & 97.72  & 97.86  & \underline{\textbf{98.11}} & \underline{\textbf{97.95}}\\
\hline
\end{tabular}
\label{tab:comparison}
\end{center}
\end{table}

\subsection{Explainability Results - Pedestrian use case}
\label{sec:exp-pedestrian-use-case}
\noindent
For this use case, explainability can be explored from two perspectives: \ac{KG} Models and \ac{RAG} models. When it comes to \ac{KG} models, the PedFeatKG ontology only activates the pedestrian features, which could provide a possible explanation for the prediction. On the other hand, if we use the PedFeatRulesKG ontology, the pedestrian features are supported by fuzzy rules, offering additional insight into the prediction.
In the second case, the pedestrian features representing the pedestrian state were queried to the retrieval module to explain why the prediction was made and whether the pedestrian will cross the road or not. 
The example depicted in \Cref{fig:explain_ped} showcases two predictions derived from JAAD videos, encompassing the prediction outcome, the frame of prediction, pedestrian features, activated fuzzy rules, and the RAG explanation. 
These examples underscore the significance of pedestrian body orientation and proximity to the road as crucial factors in explaining why pedestrians choose to cross or not cross the road. 
In the case of video 044, two fuzzy rules were activated, explaining the prediction of crossing the road due to: 1) the pedestrian is near to the road and 2) the pedestrian is oriented to the left and is walking fast.
Similarly, in the instance of video 262 where the pedestrian will not cross, one fuzzy rule was activated, indicating that pedestrians are oriented in the same direction as the vehicle. 
This explanation is further enhanced by the \ac{RAG}, which also considers the distance between the pedestrian and the ego-vehicle.

\begin{table}[h!]
\caption{Different multimedia for results visualization in the pedestrians and drivers use cases.}
\small
\begin{tblr}
{
    colspec={X[1,l] X[4,l]},
    hlines,
    vlines}
\textbf{Use Case} & \textbf{Link}  \\
Pedestrians & \url{https://www.youtube.com/playlist?list=PLAeK3AuwxenEqDvdJAk8X9Ysn5egmGvKO}\\
Drivers  & \url{https://www.youtube.com/playlist?list=PLAeK3AuwxenFsZslUIYk1CitWKAeAddgt}\\
\end{tblr}
\label{tab:driver-media}
\end{table}

\begin{figure*}[ht]
\centering
\includegraphics[width=\linewidth]{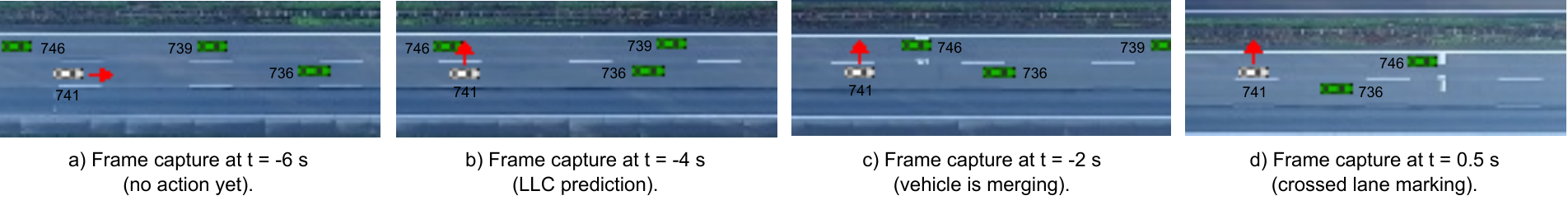}
\caption{Scene explanation through four different frames.}
\label{fig:frames_LLC_49_0741}
\end{figure*}
\begin{figure}[]
    \centering
    \includegraphics[width=\columnwidth]{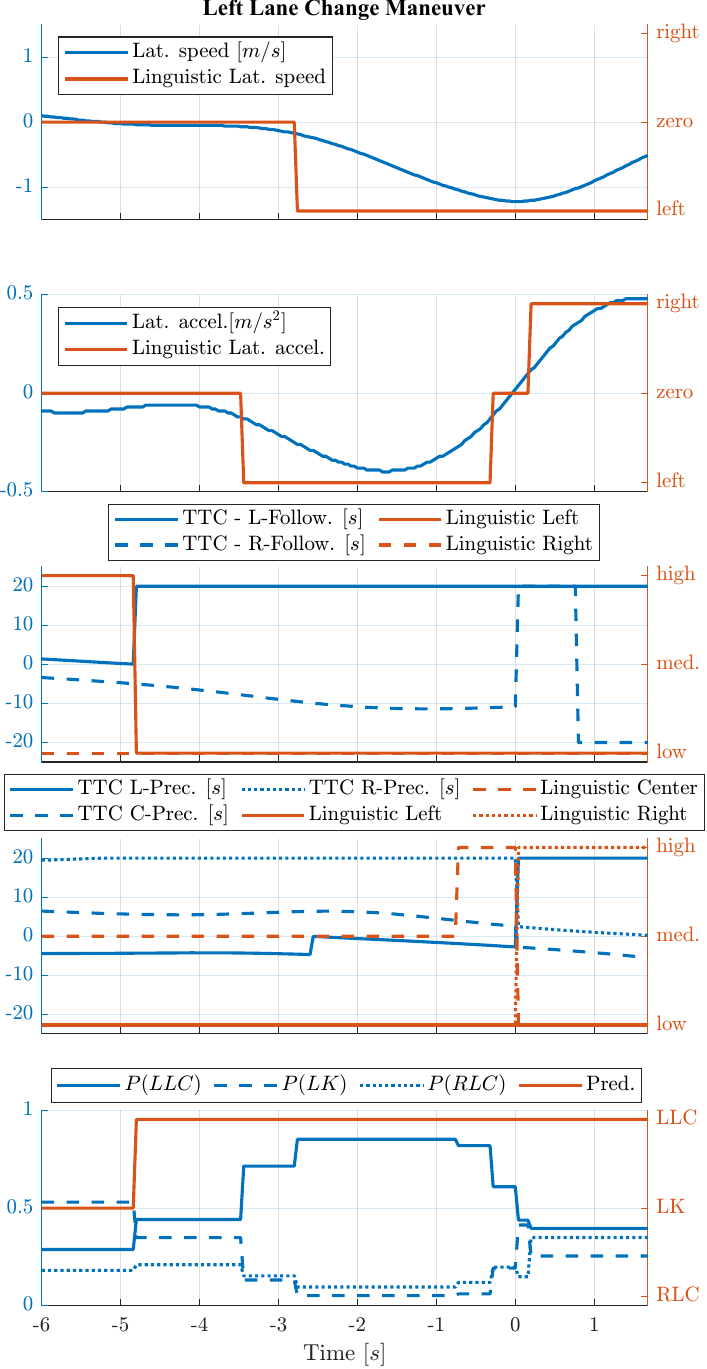}
    \caption{Temporal sequence of numerical variables and linguistic categories.}
    \label{fig:results_LLC_49_0741}
\end{figure}
\subsection{Explainability Results - Drivers use case}
\label{sec:exp-drivers-use-case}
\noindent
Regarding the explainability of the driver use case, \Cref{fig:frames_LLC_49_0741} and \Cref{fig:results_LLC_49_0741} show a scenario that involves a left lane change. The first figure captures various frames showing a white target vehicle and its green neighboring vehicles during the lane change. The second figure presents a graph detailing numerical values and linguistic data inputs fed into the \ac{KG} model for analysis of the same scenario. This includes sub-figures showing lateral velocity, lateral acceleration, and time-to-collision (TTC) with the preceding vehicle, and both following and preceding vehicles on the left and right. The final sub-figure displays the prediction probabilities throughout the scene. The aim is to demonstrate how the model employs interpretable and explainable linguistic inputs for accurate prediction.
The scene begins six seconds before the lane change as shown in \Cref{fig:frames_LLC_49_0741}, at which point the target vehicle is moving straight with zero lateral acceleration and low-risk TTC with the right preceding and following vehicles, as well as the left preceding vehicles. There is a medium risk with a center preceding vehicle and a high risk with the left following vehicle.
Using Bayesian reasoning, the model is asked to compute the probability of \textit{LLC} given the generated linguistic inputs, the same question is addressed for \textit{LK} and \textit{RLC}, and the prediction with the highest probability will be the model's prediction. 
The model uses the \ac{KGE} to get all the triples probabilities after reification as mentioned earlier in \Cref{sec:bayesian-inference-and-prediction} and \Cref{fig:pipeline}. 
During this instant, the model prediction is \textit{LK} as it has a higher probability than \textit{LLC} and \textit{RLC}.
Two seconds later, the risk associated with the left following vehicle decreases from high to low, while the medium risk with a center preceding vehicle remains. This change prompts the model to predict an \textit{LLC} represented by a red arrow pointing to the left of the vehicle.
After that, in the third captured frame represented in \Cref{fig:frames_LLC_49_0741}c. 
The target vehicle starts to accelerate in the left direction, moving with lateral velocity in the left direction as well. So, the vehicle started moving to merge and was about to change lane.
By the final frame, after the lane change by 0.5 seconds, the vehicle is merging into the left lane, accelerating right while still moving left. The preceding vehicle that was a high-risk before the lane change is now the right preceding vehicle and the left preceding vehicle has shifted positions to be directly ahead. These changes significantly affect the TTC values.
\Cref{tab:driver-media} contains links for some multimedia videos that provide results of different scenes including the scene discussed in this section.
\begin{figure*}[]
    \centering
    \includegraphics[width=\linewidth]{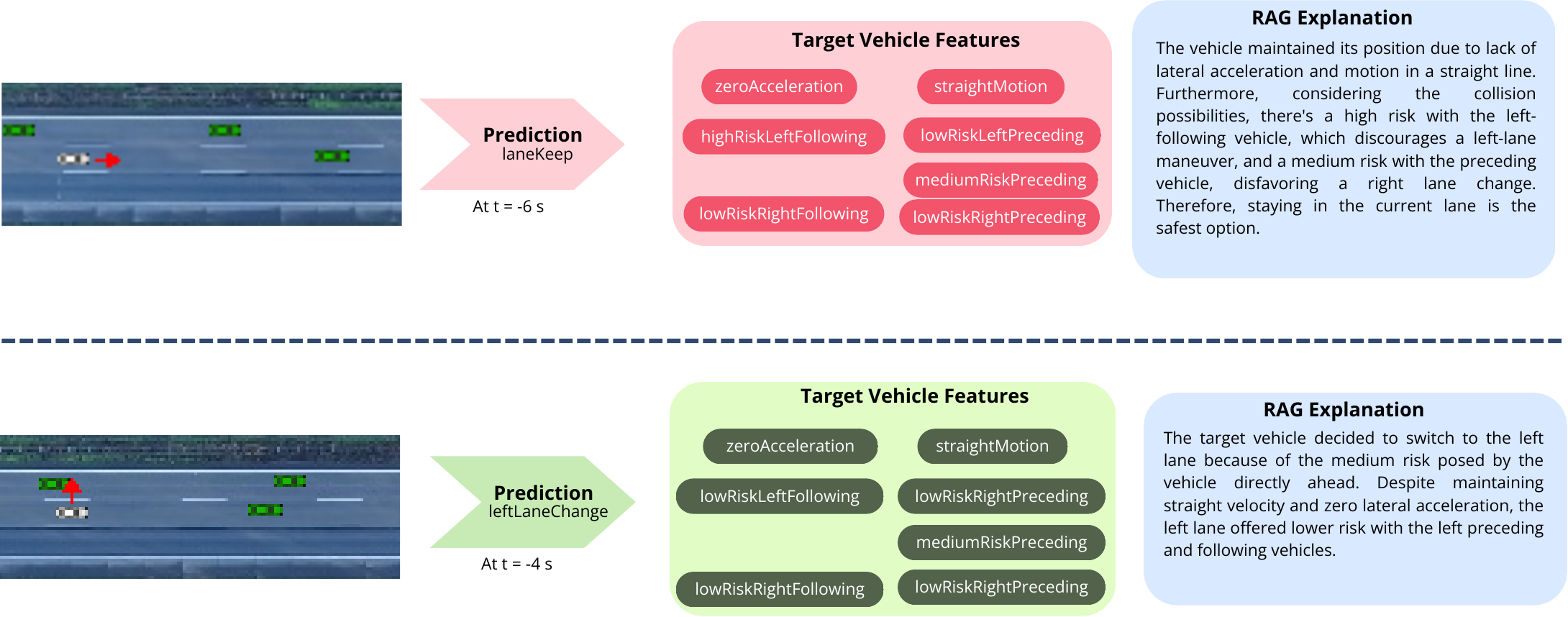}
    \caption{Examples of lane change prediction explainability based on the discussed scene from HighD dataset.}
    \label{fig:explain_vehicle}
\end{figure*}

Regarding the explainability of the driver use case using RAG, \Cref{fig:explain_vehicle} shows the RAG explanation for the first two instances in the scene described in \Cref{fig:frames_LLC_49_0741}. The model gives clear, reasonable, and precise explanations.
\section{Conclusions and future work}
\label{sec:conclusions-and-future-work}
\noindent 
In this work, a context-based road users’ behavior prediction system has been developed using Knowledge Graphs, as the main structure for representing knowledge, and Bayesian inference with graph reifications as a means to implement a fully inductive reasoning system as a downstream task. 
Two use cases have been targeted following this predictive approach: 1) pedestrian crossing actions; 2) vehicle lane change maneuvers. 
In both cases, the proposed KG-based solution provides superior performance with respect to the state of the art both in terms of anticipation and F1 metric. 
Especially relevant is the demonstrated capability for predicting road users’ behaviors in the absence of relevant kinematic clues, given the ability of the proposed system for accounting for contextual information. 
Different types of information sources have been integrated, including datasets and rules, as a proof of the capability to deal with numerical and linguistic information in a harmonized knowledge representation format using Knowledge Graphs. 
This feature endows the system with the capacity to incorporate human knowledge in the form of linguistic descriptions representing experience and/or rules.  
Finally, explainable descriptions of the behavioral predictions have been implemented using Retrieval Augmented Generation Techniques (RAG), as a means to combine the reasoning ability of Knowledge Graphs and the expressive capacity of Large Language Models.
Despite the progress exhibited in the current work, a number of improvements are envisaged with a view to extending and testing the predictive capabilities in new use cases, such as: 1) near-miss (or crash) lane change maneuvers, and 2) occluded children on urban scenarios. Similarly, further research is necessary for getting to understand road users’ behaviors in a more holistic manner, especially in cross-cultural settings. For that purpose, new data will be gathered in regions of the world with different social rules, such as the MENA (Middle East and North of Africa) region, South-East Asia, and Latin America.
Finally, the proposed predictive system will be integrated with the behavior planner of an Autonomous Vehicle in order to make AVs to behave in a more human-like fashion. 

\section*{Acknowledgment}
\noindent This research has been funded by the HEIDI project of the European Commission under Grant Agreement: 101069538.

\newpage

\bibliographystyle{IEEEtran}
\bibliography{article_v2}

\begin{thebibliography}{10}
\providecommand{\url}[1]{#1}
\csname url@samestyle\endcsname
\providecommand{\newblock}{\relax}
\providecommand{\bibinfo}[2]{#2}
\providecommand{\BIBentrySTDinterwordspacing}{\spaceskip=0pt\relax}
\providecommand{\BIBentryALTinterwordstretchfactor}{4}
\providecommand{\BIBentryALTinterwordspacing}{\spaceskip=\fontdimen2\font plus
\BIBentryALTinterwordstretchfactor\fontdimen3\font minus \fontdimen4\font\relax}
\providecommand{\BIBforeignlanguage}[2]{{%
\expandafter\ifx\csname l@#1\endcsname\relax
\typeout{** WARNING: IEEEtran.bst: No hyphenation pattern has been}%
\typeout{** loaded for the language `#1'. Using the pattern for}%
\typeout{** the default language instead.}%
\else
\language=\csname l@#1\endcsname
\fi
#2}}
\providecommand{\BIBdecl}{\relax}
\BIBdecl

\bibitem{WHO}
W.~H. Organization, ``Global status report on road safety,'' 2023.

\bibitem{Freya}
F.~Slootmans, ``European road safety observatory. technical report. european comission,'' 2022.

\bibitem{Stewart}
T.~Stewart, ``Overview of motor vehicle traffic crashes,'' 2021.

\bibitem{Tsotsos1}
J.~K.~T. Iuliia~Kotseruba, Amir~Rasouli, ``Benchmark for evaluating pedestrian action prediction,'' \emph{Technical report}, 2021.

\bibitem{Pool}
D.~M.~G. E.~A. I.~Pool, J. F. P.~Kooij, ``Crafted vs. learned representations in predictive models: A case study on cyclist path prediction,'' \emph{IEEE Transactions on Intelligent Vehicles}, 2021.

\bibitem{Izquierdo1}
R.~I. et~al, ``Vehicle trajectory prediction on highways using bird eye view representations and deep learning,'' \emph{Applied Intelligence}, pp. 1--19, 2022.

\bibitem{Schulz}
R.~S. T.~Schulz, ``A controlled interactive multiple model filter for combined pedestrian intention recognition and path prediction,'' \emph{IEEE Intelligent Transportation Systems Conference}, pp. 173--178, 2015.

\bibitem{Su}
R.~I. et~al, ``A. su, k. muelling, j. dolan, p. palanisamy, p. mudalige,'' \emph{IEEE Intelligent Vehicles Symposium}, pp. 1412--1417, 2018.

\bibitem{Benterki}
A.~Benterki, M.~Boukhnifer, V.~Judalet, and M.~Choubeila, ``Prediction of surrounding vehicles lane change intention using machine learning,'' \emph{IEEE International Conference on Intelligent Data Acquisition and Advanced Computing Systems: Technology and Applications}, pp. 839--843, 2019.

\bibitem{Izquierdo2}
R.~Izquierdo, A.~Quintanar, I.~Parra, D.~Fernández-Llorca, and M.~A. Sotelo, ``Experimental validation of lane-change intention prediction methodologies based on cnn and lstm,'' \emph{IEEE Intelligent Transportation Systems Conference}, pp. 3657--3662, 2019.

\bibitem{Izquierdo3}
R.~Izquierdo, A.~Quintanar, I.~Parra-Alonso, D.~F. Llorca, and M.~A. Sotelo, ``The prevention – a novel benchmark for prediction of vehicles intentions,'' \emph{IEEE Intelligent Transportation Systems Conference}, 2019.

\bibitem{Laimona}
O.~Laimona, M.~A. Manzour, O.~M. Shehata, and E.~I. Morgan, ``Implementation and evaluation of an enhanced intention prediction algorithm for lane-changing scenarios on highway roads,'' \emph{2nd Novel Intelligent and Leading Emerging Sciences Conference}, pp. 128--133, 2020.

\bibitem{Xue}
J.~L. Q.~Xue, Y.~Xing, ``An integrated lane change prediction model incorporating traffic context based on trajectory data,'' \emph{Transportation research part C: emerging technologies}, vol. 141, 2022.

\bibitem{Krajweski}
R.~K. et~al, ``Experimental insights towards explainable and interpretable pedestrian crossing prediction,'' 2018.

\bibitem{Gao}
K.~Gao, X.~Li, B.~Chen, L.~Hu, J.~Liu, R.~Du, and Y.~Li, ``Dual transformer based prediction for lane change intentions and trajectories in mixed traffic environment,'' \emph{IEEE Transactions on Intelligent Transportation Systems}, 2023.

\bibitem{Tsotsos2}
J.~K.~T. Iuliia~Kotseruba, Amir~Rasouli, ``Do they want to cross? understanding pedestrian intention for behavior prediction,'' \emph{IEEE Intelligent Vehicles Symposium (IV)}, pp. 1688--1693, 2020.

\bibitem{Lorenzo}
M.~A.~S. Javier~Lorenzo, Ignacio~Parra, ``Capformer: Pedestrian crossing action prediction using transformer,'' \emph{Sensors}, 2021.

\bibitem{Tran}
D.~T. et~al, ``Learning spatiotemporal features with 3d convolutional networks,'' 2015.

\bibitem{Tsotsos3}
J.~K.~T. Amir~Rasouli, Iuliia~Kotseruba, ``Pedestrian action anticipation using contextual feature fusion in stacked rnns,'' 2020.

\bibitem{Xingjian}
X.~et~al, ``Convolutional lstm network: A machine learning approach for precipitation nowcasting,'' \emph{Advances in Neural Information Processing Systems}, vol.~28, 2015.

\bibitem{Achaji}
L.~A. et~al, ``Analysis over vision-based models for pedestrian action anticipation,'' 2023.

\bibitem{Muscholl}
N.~M. et~al, ``Emidas: explainable social interaction-based pedestrian intention detection across street,'' \emph{Annual ACM Symposium on Applied Computing}, 2021.

\bibitem{Yi}
K.~Yi, J.~Wu, C.~Gan, A.~Torralba, P.~Kohli, and J.~B. Tenenbaum, ``Neural-symbolic vqa: Disentangling reasoning from vision and language understanding,'' 2019.

\bibitem{Hogan}
A.~H. et~al, ``Knowledge graphs,'' 2020.

\bibitem{Martin}
A.~Martin, K.~Hinkelmann, H.-G. Fill, A.~Gerber, D.~Lenat, R.~Stolle, and F.~van Harmelen, ``An evaluation of knowledge graph embeddings for autonomous driving data: Experience and practice,'' \emph{Proceedings of the AAAI 2020 Spring Symposium on Combining Machine Learning and Knowledge Engineering in Practice}, 2020.

\bibitem{kgdef}
C.~Peng, F.~Xia, M.~Naseriparsa, and F.~Osborne, ``Knowledge graphs: Opportunities and challenges,'' 2023.

\bibitem{kgedef}
S.~Choudhary, T.~Luthra, A.~Mittal, and R.~Singh, ``A survey of knowledge graph embedding and their applications,'' 2021.

\bibitem{transE}
\BIBentryALTinterwordspacing
A.~Bordes, N.~Usunier, A.~Garcia-Duran, J.~Weston, and O.~Yakhnenko, ``Translating embeddings for modeling multi-relational data,'' in \emph{Advances in Neural Information Processing Systems}, C.~Burges, L.~Bottou, M.~Welling, Z.~Ghahramani, and K.~Weinberger, Eds., vol.~26.\hskip 1em plus 0.5em minus 0.4em\relax Curran Associates, Inc., 2013. [Online]. Available: \url{https://proceedings.neurips.cc/paper_files/paper/2013/file/1cecc7a77928ca8133fa24680a88d2f9-Paper.pdf}
\BIBentrySTDinterwordspacing

\bibitem{complex}
\BIBentryALTinterwordspacing
T.~Trouillon, J.~Welbl, S.~Riedel, {\'{E}}.~Gaussier, and G.~Bouchard, ``Complex embeddings for simple link prediction,'' \emph{CoRR}, vol. abs/1606.06357, 2016. [Online]. Available: \url{http://arxiv.org/abs/1606.06357}
\BIBentrySTDinterwordspacing

\bibitem{psi}
T.~Chen, T.~Jing, R.~Tian, Y.~Chen, J.~Domeyer, H.~Toyoda, R.~Sherony, and Z.~Ding, ``Psi: A pedestrian behavior dataset for socially intelligent autonomous car,'' \emph{arXiv preprint arXiv:2112.02604}, 2021.

\bibitem{kg_ped_predictor}
A.~N. Melo, L.~F. Herrera-Quintero, C.~Salinas, and M.~A. Sotelo, ``Knowledge-based explainable pedestrian behavior predictor,'' 2024.

\bibitem{highDdataset}
R.~Krajewski, J.~Bock, L.~Kloeker, and L.~Eckstein, ``The highd dataset: A drone dataset of naturalistic vehicle trajectories on german highways for validation of highly automated driving systems,'' in \emph{2018 21st International Conference on Intelligent Transportation Systems (ITSC)}, 2018, pp. 2118--2125.

\bibitem{manzour2023vehicle}
M.~Manzour, A.~Ballardini, R.~Izquierdo, and M.~Sotelo, ``Vehicle lane change prediction based on knowledge graph embeddings and bayesian inference,'' \emph{arXiv preprint arXiv:2312.06336}, 2023.

\bibitem{saffarzadeh2013general}
M.~Saffarzadeh, N.~Nadimi, S.~Naseralavi, and A.~R. Mamdoohi, ``A general formulation for time-to-collision safety indicator,'' in \emph{Proceedings of the Institution of Civil Engineers-Transport}, vol. 166, no.~5.\hskip 1em plus 0.5em minus 0.4em\relax Thomas Telford Ltd, 2013, pp. 294--304.

\bibitem{ramezani2020comparing}
E.~RAMEZANI-KHANSARI, F.~M. NEJAD, and S.~MOOGEH, ``Comparing time to collision and time headway as safety criteria,'' \emph{Pamukkale {\"U}niversitesi M{\"u}hendislik Bilimleri Dergisi}, vol.~27, no.~6, pp. 669--675, 2020.

\bibitem{ampligraph}
\BIBentryALTinterwordspacing
L.~Costabello, A.~Bernardi, A.~Janik, S.~Pai, C.~L. Van, R.~McGrath, N.~McCarthy, and P.~Tabacof, ``{AmpliGraph: a Library for Representation Learning on Knowledge Graphs},'' Mar. 2019. [Online]. Available: \url{https://doi.org/10.5281/zenodo.2595043}
\BIBentrySTDinterwordspacing

\bibitem{ivturs}
J.~Sanz, A.~Fernandez, H.~Bustince, and F.~Herrera, ``Ivturs: a linguistic fuzzy rule-based classification system based on a new interval-valued fuzzy reasoning method with tuning and rule selection,'' \emph{IEEE Transactions on Fuzzy Systems}, vol.~21, no.~3, pp. 399--411, 2013.

\bibitem{farchd}
J.~Alcala-Fdez, R.~Alcala, and F.~Herrera, ``A fuzzy association rule-based classification model for high-dimensional problems with genetic rule selection and lateral tuning,'' \emph{IEEE Transactions on Fuzzy Systems}, vol.~19, no.~5, pp. 857--872, 2011.

\bibitem{rulesdef}
H.~Ishibuchi and T.~Nakashima, ``Effect of rule weights in fuzzy rule-based classification systems,'' \emph{IEEE Transactions on Fuzzy Systems}, vol.~9, no.~4, pp. 506--515, 2001.

\bibitem{lewis2020retrieval}
P.~Lewis, E.~Perez, A.~Piktus, F.~Petroni, V.~Karpukhin, N.~Goyal, H.~K{\"u}ttler, M.~Lewis, W.-t. Yih, T.~Rockt{\"a}schel \emph{et~al.}, ``Retrieval-augmented generation for knowledge-intensive nlp tasks,'' \emph{Advances in Neural Information Processing Systems}, vol.~33, pp. 9459--9474, 2020.

\bibitem{melo2023experimental}
A.~N. Melo, C.~Salinas, and M.~A. Sotelo, ``Experimental insights towards explainable and interpretable pedestrian crossing prediction,'' 2023.

\bibitem{yolopv2}
C.~Han, Q.~Zhao, S.~Zhang, Y.~Chen, Z.~Zhang, and J.~Yuan, ``Yolopv2: Better, faster, stronger for panoptic driving perception,'' 2022.

\bibitem{pedrecnet}
\BIBentryALTinterwordspacing
D.~Burgermeister and C.~Curio, ``Pedrecnet: Multi-task deep neural network for full 3d human pose and orientation estimation,'' in \emph{2022 {IEEE} Intelligent Vehicles Symposium, {IV} 2022, Aachen, Germany,June 4-9, 2022}.\hskip 1em plus 0.5em minus 0.4em\relax {IEEE}, 2022, pp. 441--448. [Online]. Available: \url{https://doi.org/10.1109/IV51971.2022.9827202}
\BIBentrySTDinterwordspacing

\bibitem{cd3}
\BIBentryALTinterwordspacing
D.~Tran, L.~D. Bourdev, R.~Fergus, L.~Torresani, and M.~Paluri, ``{C3D:} generic features for video analysis,'' \emph{CoRR}, vol. abs/1412.0767, 2014. [Online]. Available: \url{http://arxiv.org/abs/1412.0767}
\BIBentrySTDinterwordspacing

\bibitem{benchcrossing}
I.~Kotseruba, A.~Rasouli, and J.~K. Tsotsos, ``Benchmark for evaluating pedestrian action prediction,'' in \emph{2021 IEEE Winter Conference on Applications of Computer Vision (WACV)}, 2021, pp. 1257--1267.

\bibitem{xue2022integrated}
Q.~Xue, Y.~Xing, and J.~Lu, ``An integrated lane change prediction model incorporating traffic context based on trajectory data,'' \emph{Transportation research part C: emerging technologies}, vol. 141, p. 103738, 2022.

\bibitem{gao2023dual}
K.~Gao, X.~Li, B.~Chen, L.~Hu, J.~Liu, R.~Du, and Y.~Li, ``Dual transformer based prediction for lane change intentions and trajectories in mixed traffic environment,'' \emph{IEEE Transactions on Intelligent Transportation Systems}, 2023.

\end{thebibliography}

\begin{IEEEbiography}[{\includegraphics
[width=1in,height=1.25in,clip,
keepaspectratio]{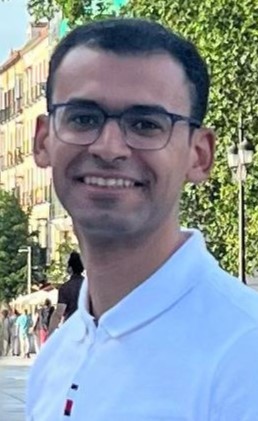}}]
{Mohamed Manzour Hussien}
obtained his bachelor's degree in mechatronics from the German University in Cairo (GUC) in 2019. During his undergraduate studies, he had the opportunity to conduct his bachelor's thesis in the field of machine learning at the IFS (Institut für Schienenfahrzeuge) institute in Germany. After that, he worked as a lecturer assistant in the GUC till 2022. During this period, he completed his master's degree in the field of Intelligent Transportation Systems (ITS) at the Multi-Robot Systems (MRS) research group in the GUC in 2022. He focused on pedestrian behavior prediction. In 2023, he started his Ph.D. journey in the field of ITS at the INVETT (INtelligent VEhicles and Traffic Technologies) research group, University of Alcala, Spain. He is focusing on vehicle behavior prediction. In 2023, he won the IEEE ITSS student competition in the track of Driver Decision Prediction.
\end{IEEEbiography}

\begin{IEEEbiography}[{\includegraphics
[width=1in,height=1.25in,clip,
keepaspectratio]{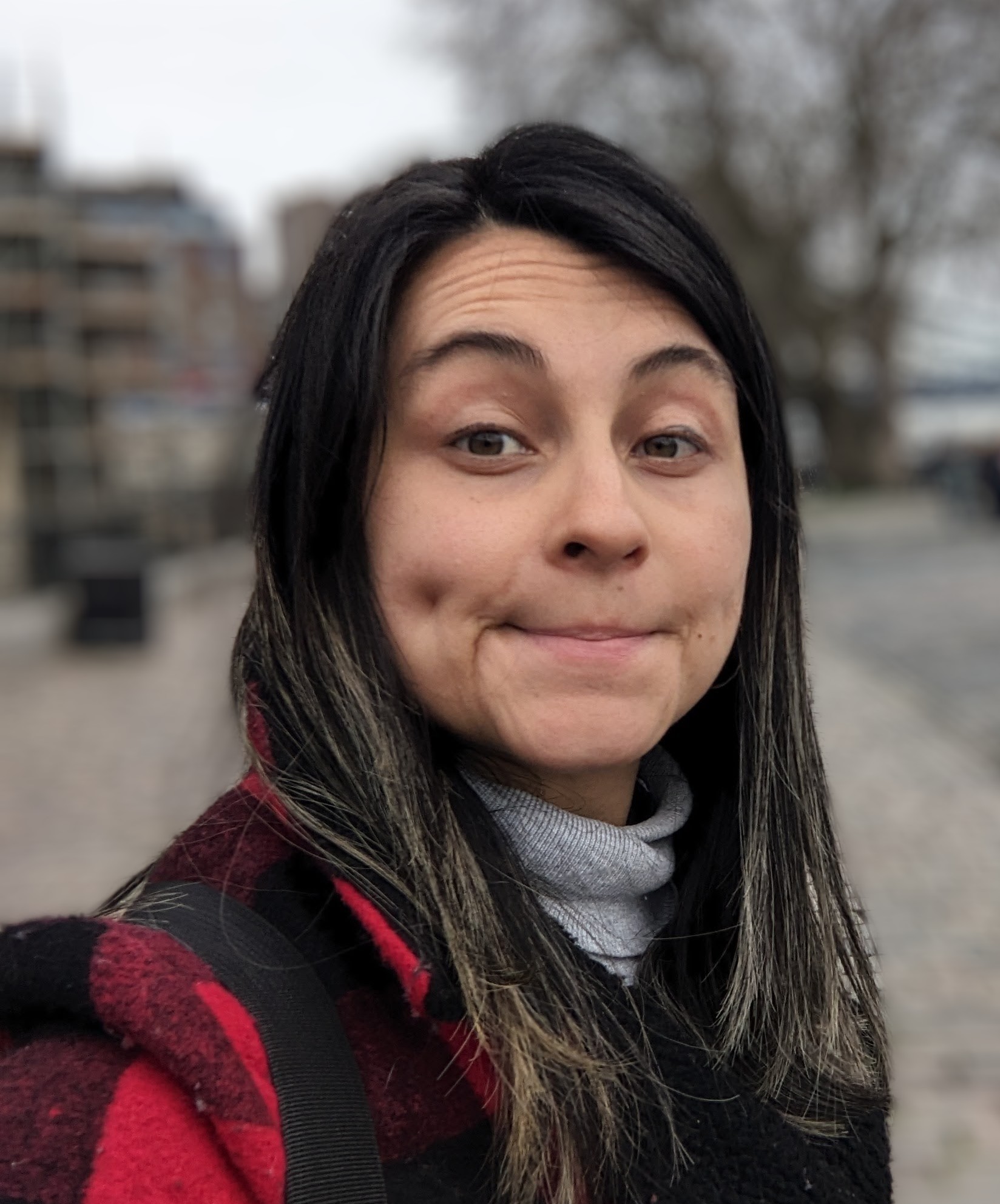}}]
{Angie Nataly Melo }
obtained the Bacherlor's Degree in Informatics in 2013 from University Catolica from Colombia and the Master's Degree in Intelligent Transportation Systems in 2016 from Czech Technical University in Prague. She started her work in the INVETT Research Group in September 2022, where she is currently pursing the PhD degree in Information and Communications Technologies with the Computer Engineering Department, developing explainable prediction systems in the context of pedestrian behavior.
\end{IEEEbiography}

\begin{IEEEbiography}[{\includegraphics
[width=1in,height=1.25in,clip,
keepaspectratio]{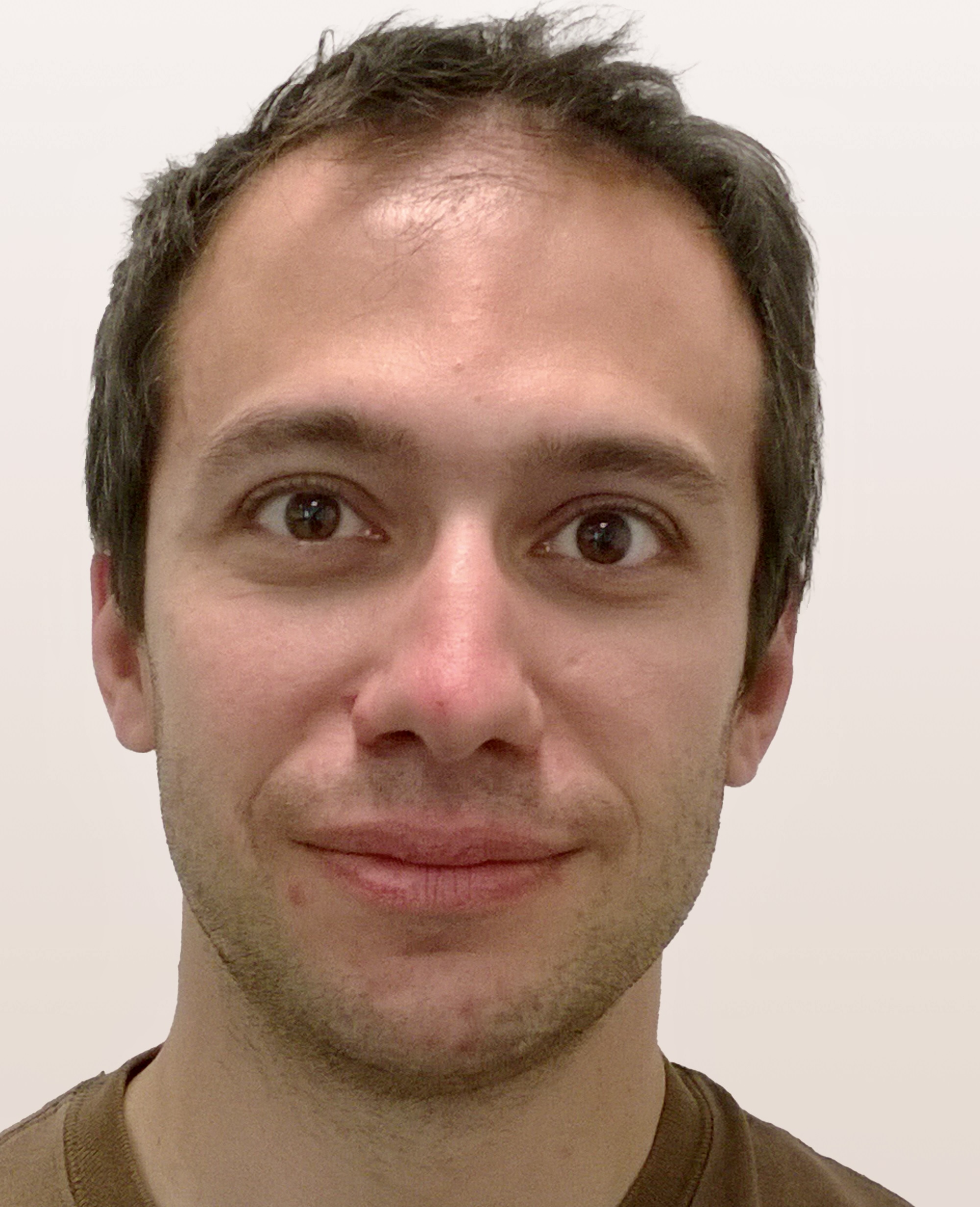}}]
{Augusto Luis Ballardini}
was born in Buenos Aires, Argentina, in 1984. He completed his M.Sc. and Ph.D. degrees in Computer Science from the Università degli Studi di Milano - Bicocca, Italy, in 2012 and 2017 respectively. Following his post-doctoral activities in the IRALAB Research Group for two years, Dr. Ballardini joined the INVETT Research Group at the Universidad de Alcalá, Spain, in 2019. During his time at INVETT, he was awarded a Marie Skłodowska-Curie Actions research grant and a research grant within the Maria Zambrano/NextGenerationEU project from the Spanish Ministry of Science, Innovation, and Universities. His research focuses on developing advanced systems for autonomous vehicle localization and data fusion, using heterogeneous data sources such as digital maps, LiDAR, and image data, combined with cutting-edge computer vision and machine learning algorithms. 
\end{IEEEbiography}

\begin{IEEEbiography}[{\includegraphics
[width=1in,height=1.25in,clip,
keepaspectratio]{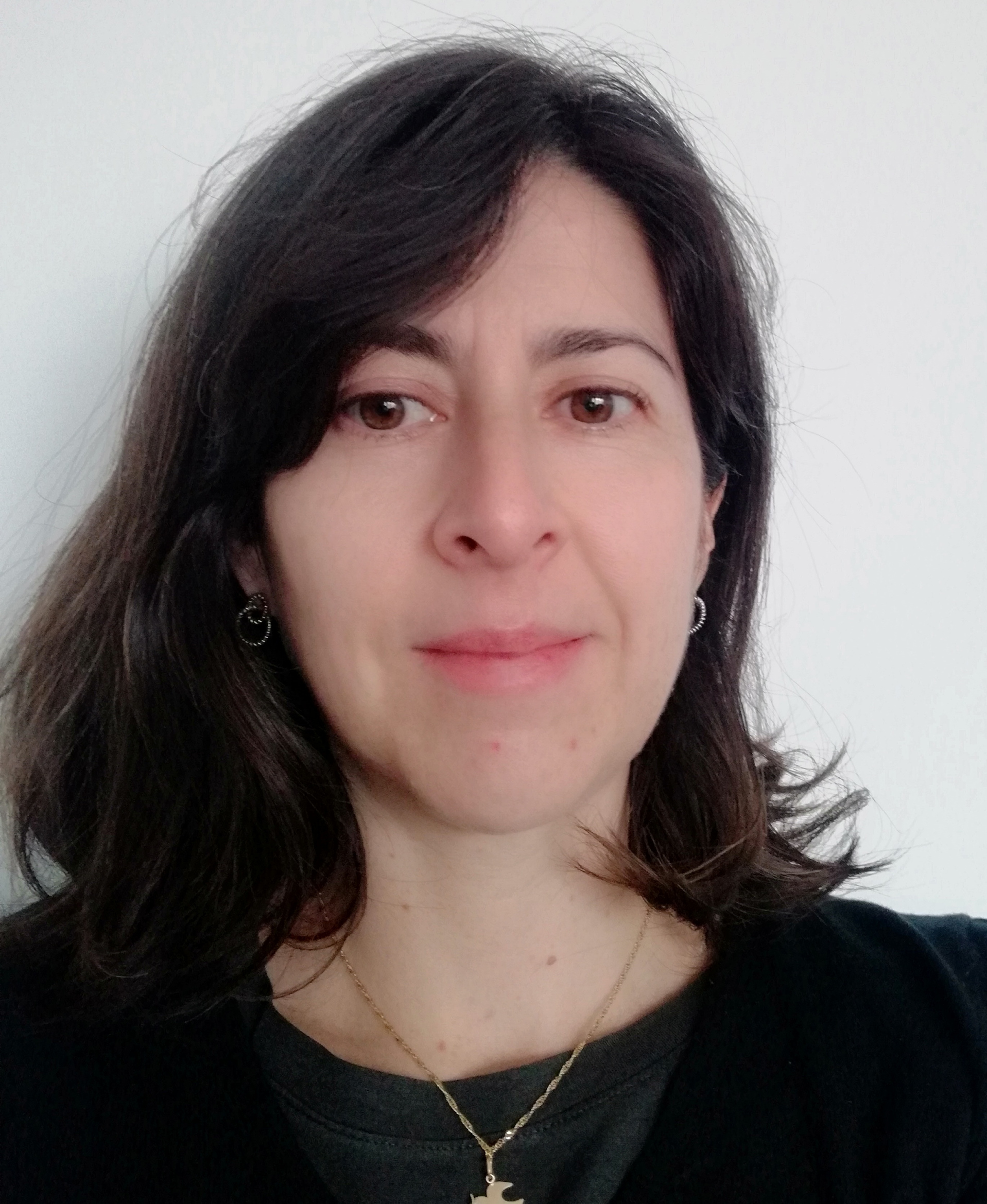}}]
{Carlota Salinas Maldonado}
earned her Ph.D. degree in engineering and automatics from the Universidad Complutense de Madrid in 2015. She is currently an assistant professor at the Computer Engineering Department, University of Alcalá, Alcalá de Henares, Madrid, 28801, Spain. Her research interests include autonomous vehicle navigation, data fusion systems, lidar, computer vision, and machine learning algorithms.
\end{IEEEbiography}

\begin{IEEEbiography}[{\includegraphics
[width=1in,height=1.25in,clip,
keepaspectratio]{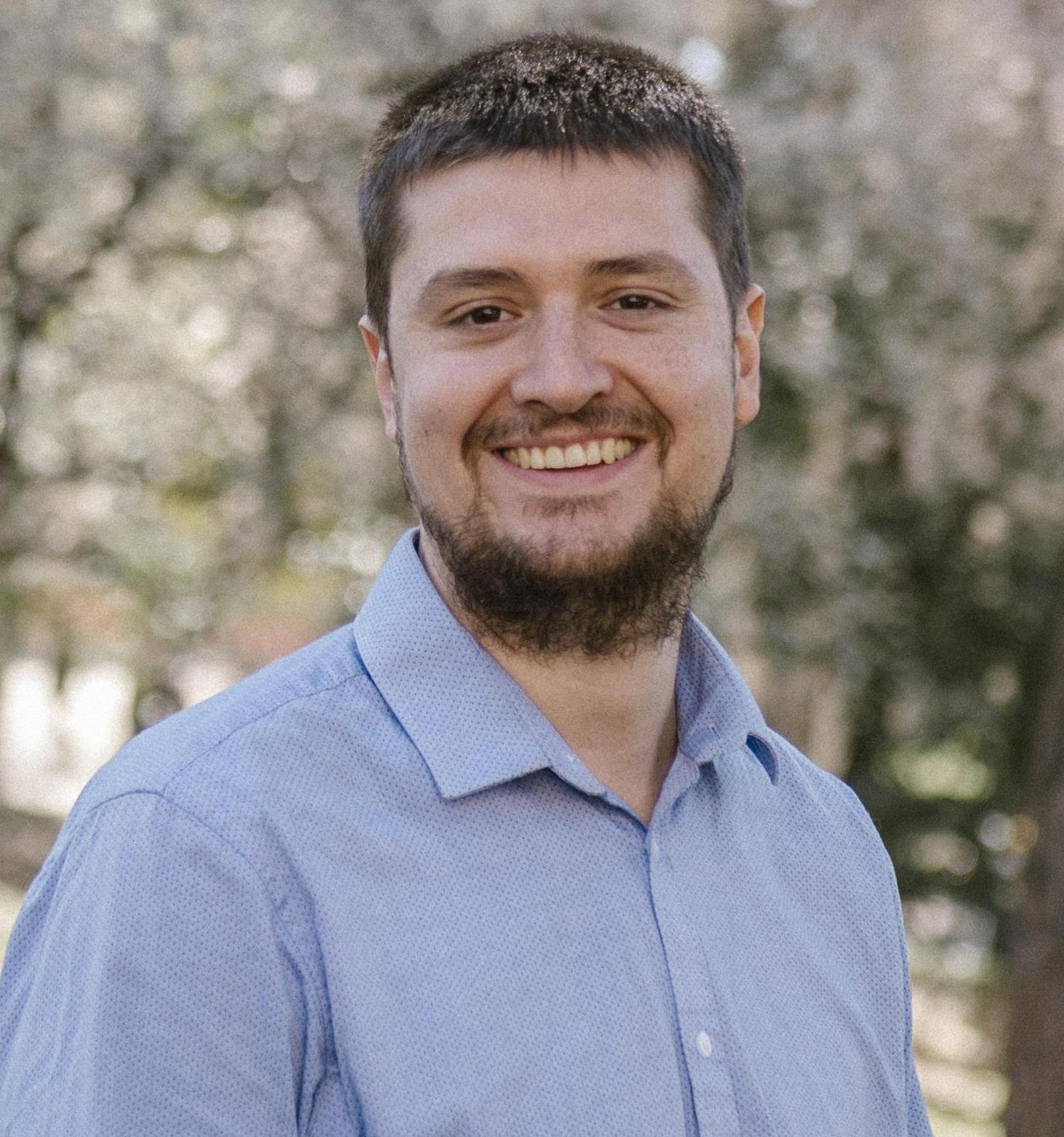}}]
{Rubén Izquierdo}
received the Bachelor’s degree in electronics and industrial automation engineering in 2014, the M.S. in industrial engineering in 2016, and the Ph.D. degree in information and communication technologies in 2020 from the University of Alcalá (UAH). He is currently Assistant Professor at the Department of Computer Engineering of the UAH. His research interest is focused on the prediction of vehicle behaviors and control algorithms for highly automated and cooperative vehicles. His work has developed a predictive ACC and AES system for cut-in collision avoidance successfully tested in Euro NCAP tests. He was awarded with the Best Ph.D. thesis on Intelligent Transportation Systems by the Spanish Chapter of the ITSS in 2022, the outstanding award for his Ph.D. thesis by the UAH in 2021. He also received the XIII Prize from the Social Council of the UAH to the University-Society Knowledge Transfer in 2018 and the Prize to the Best Team with Full Automation in GCDC 2016.
\end{IEEEbiography}

\begin{IEEEbiography}[{\includegraphics
[width=1in,height=1.25in,clip,
keepaspectratio]{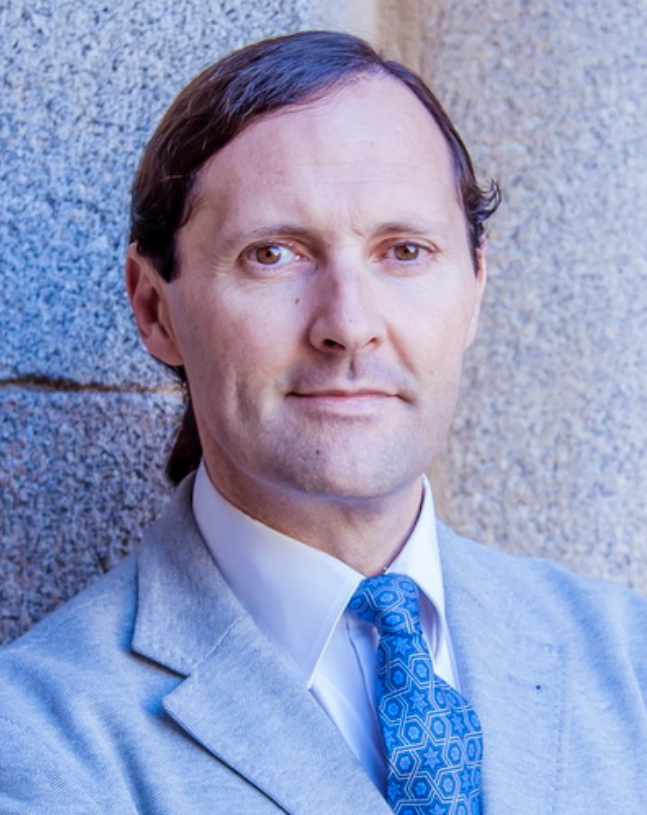}}]
{Miguel Ángel Sotelo}
received the degree in Electrical Engineering in 1996 from the Technical University of Madrid, the Ph.D. degree in Electrical Engineering in 2001 from the University of Alcalá (Alcalá de Henares, Madrid), Spain, and the Master in Business Administration (MBA) from the European Business School in 2008. He is currently a Full Professor at the Department of Computer Engineering of the University of Alcalá (UAH). His research interests include Self-driving cars, Prediction Systems, and Traffic Technologies. He is author of more than 300 publications in journals, conferences, and book chapters. He has been recipient of the Best Research Award in the domain of Automotive and Vehicle Applications in Spain in 2002 and 2009, and the 3M Foundation Awards in the category of eSafety in 2004 and 2009. Miguel Ángel Sotelo has served as Project Evaluator, Rapporteur, and Reviewer for the European Commission in the field of ICT for Intelligent Vehicles and Cooperative Systems in FP6 and FP7. He was Editor-in-Chief of the IEEE Intelligent Transportation Systems Magazine (2014-2016), Associate Editor of IEEE Transactions on Intelligent Transportation Systems (2008-2014), member of the Steering Committee of the IEEE Transactions on Intelligent Vehicles (since 2015), and a member of the Editorial Board of The Open Transportation Journal (2006-2015). He has served as General Chair of the 2012 IEEE Intelligent Vehicles Symposium (IV’2012) that was held in Alcalá de Henares (Spain) in June 2012. He was recipient of the IEEE ITS Outstanding Research Award in 2022, the IEEE ITS Outstanding Application Award in 2013, and the Prize to the Best Team with Full Automation in GCDC 2016. He is a Former President of the IEEE Intelligent Transportation Systems Society.
\end{IEEEbiography}

\end{document}